\documentclass[lettersize,journal]{IEEEtran}
\usepackage{amsmath,amsfonts}
\usepackage{algorithmic}
\usepackage{algorithm}
\usepackage{array}
\usepackage{subcaption}
\usepackage{textcomp}
\usepackage{stfloats}
\usepackage{url}
\usepackage{verbatim}
\usepackage{graphicx}
\usepackage{cite}
\usepackage{multirow}
\usepackage{threeparttable}
\usepackage{booktabs}
\usepackage[colorlinks=true, linkcolor=blue, urlcolor=blue, citecolor=blue]{hyperref}
\usepackage{bbold}
\usepackage{makecell}
\usepackage{listings}
\usepackage{color}
\usepackage{xcolor}

\newif\iffinal

\finalfalse

\iffinal
  \newcommand{\todo}[1]{}
\else
  \newcommand{\todo}[1]{{\textcolor{red}{#1}}}
  \newcommand{\edit}[1]{{\textcolor{black}{#1}}}
\fi

\newcommand{\figSigmoidTree}{
    \begin{figure}[t]
        \centering
        \includegraphics[width=0.35\textwidth]{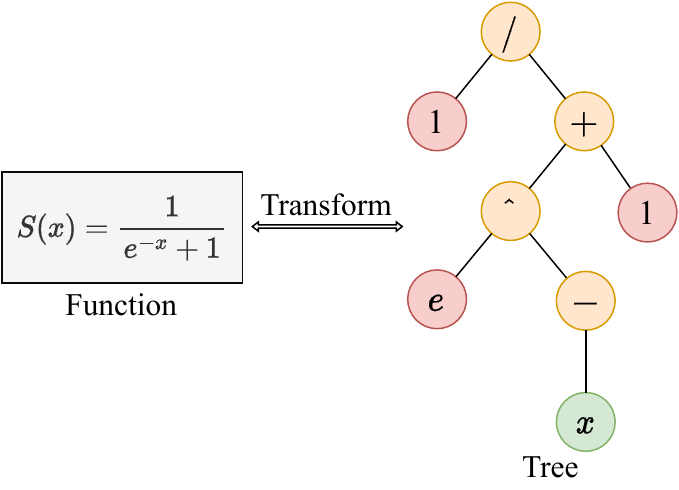}
        \caption{An example of a tree. In TGP, a computational process is represented as a tree structure. The tree shown in the figure illustrates the sigmoid function. In the figure, the orange nodes represent function nodes, the red nodes represent constant nodes, and the green nodes represent variable nodes. 
    }
        \label{fig:sigmoid_tree}
    \end{figure}
}

\newcommand{\figPopEncoding}{
    \begin{figure*}[tbp]
        \centering
        \includegraphics[width=0.9\textwidth]{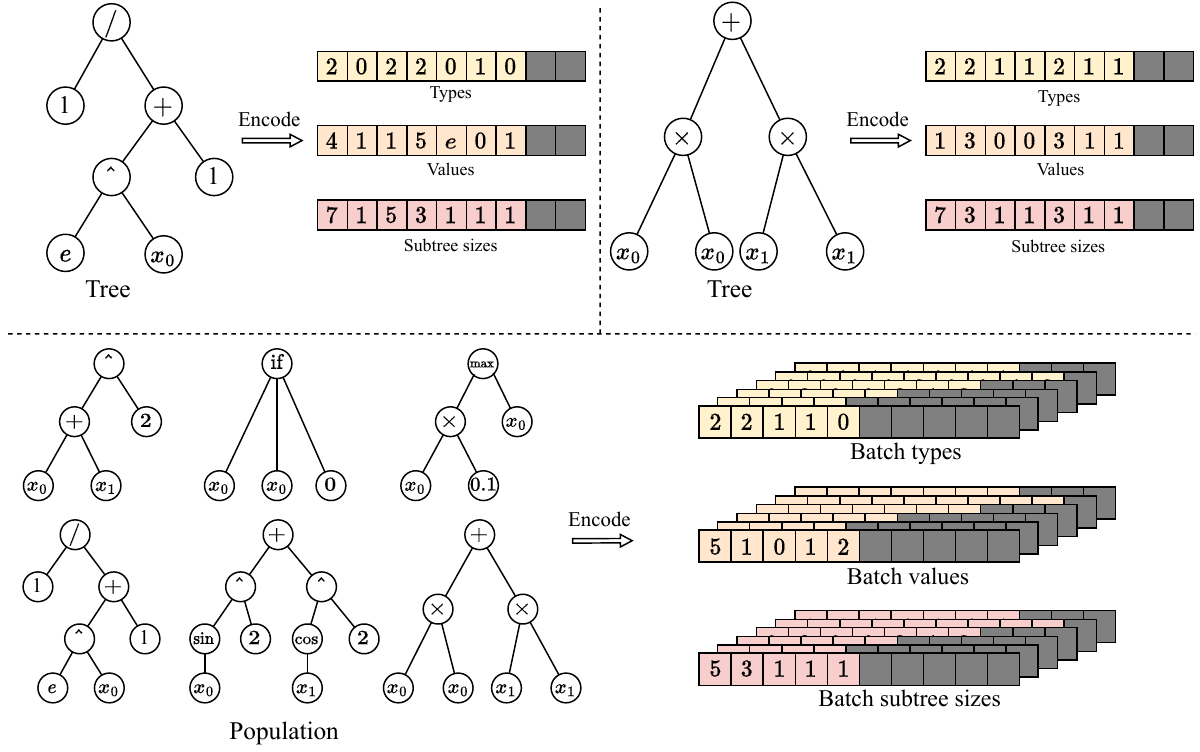}
        \caption{Illustration of the tree encoding process.
        In tensorized encoding, a tree is encoded into three tensors: types, values, and subtree sizes. We use \texttt{NaN} padding to ensure that these tensors reach a uniform maximum size, allowing trees of different structures to be encoded into tensors of the same shape. This enables the encoding of an entire population of trees into three batched tensors.
        }
        \label{fig:pop_encoding}
    \end{figure*}
}

\newcommand{\figEaOperation}{
    \begin{figure*}[tbp]
        \centering
        \includegraphics[width=0.95\textwidth]{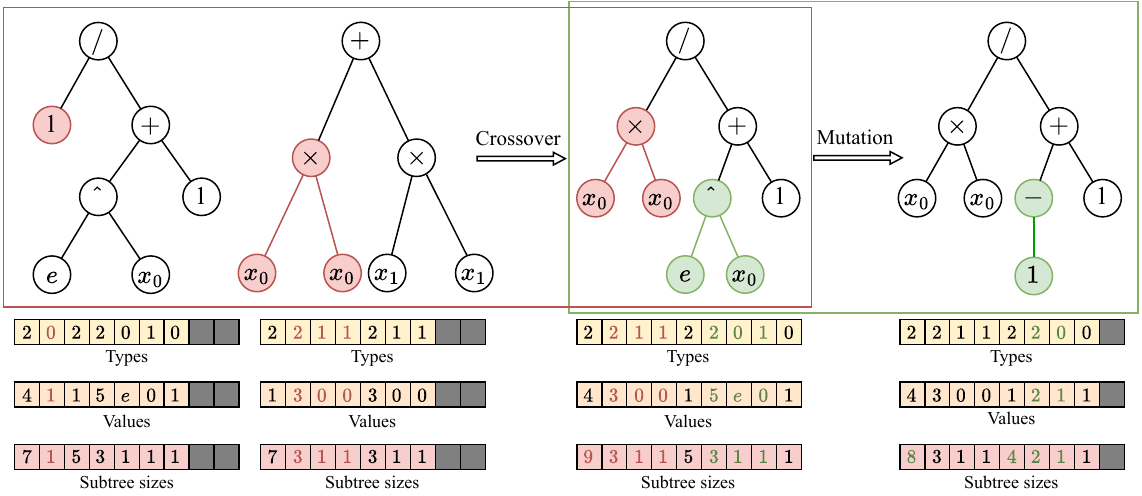}
        \caption{An illustration of genetic operations in TGP. The upper part depicts the structural modifications of trees. The red box highlights the crossover operation, where two parent trees exchange subtrees to generate a new tree. The green box illustrates the mutation process, where a subtree of a given tree is replaced with a newly generated subtree. The lower part of the figure demonstrates the corresponding transformations in the tensor representation of trees for both crossover and mutation operations.}
        \label{fig:ea_operation}
    \end{figure*}
}

\newcommand{\figCudaBlocks}{
    \begin{figure*}[bp]
        \centering
        \includegraphics[width=0.9\textwidth]{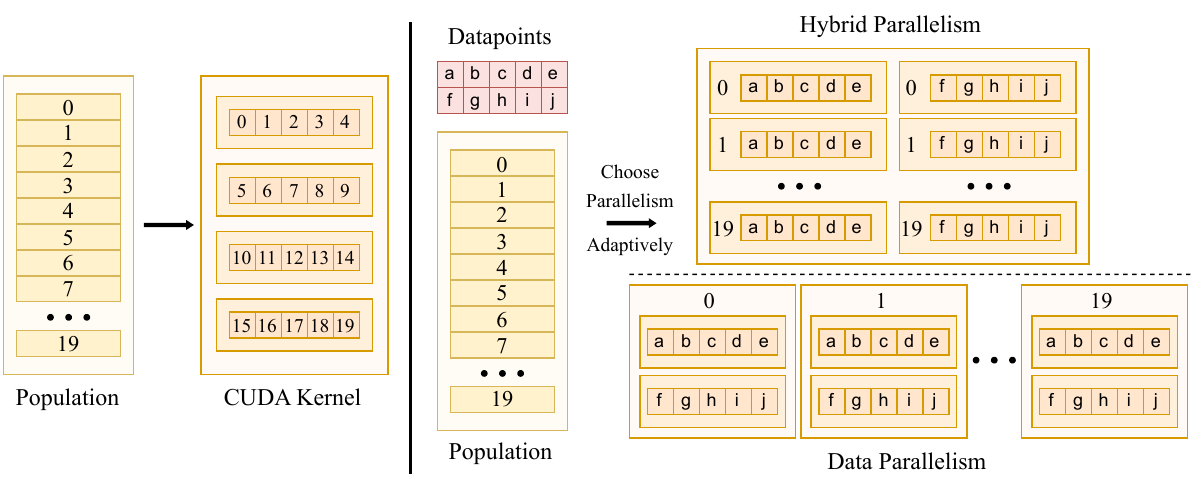}
        \caption{
        Illustration of CUDA-based parallelism in EvoGP. \textbf{Left}: Population parallelism for generation, crossover, mutation and inference kernel, where each individual is assigned to a separate thread. \textbf{Right}: SR fitness evaluation, with hybrid parallelism (upper) computing all trees in one kernel launch, while data parallelism (lower) processes each tree separately across multiple launches. 
        The system adaptively switches between the two modes based on dataset size.
        }
        \label{fig:cuda_blocks}
    \end{figure*}
}

\newcommand{\figStructure}{
    \begin{figure}[htb]
        \centering
        \includegraphics[width=0.45\textwidth]{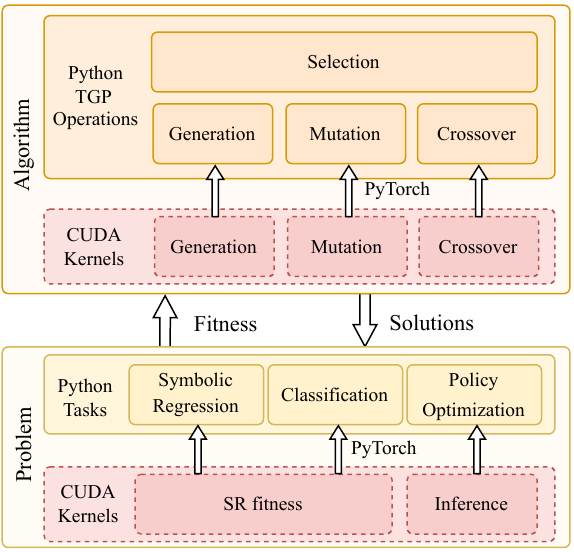}
        \caption{Illustration of the EvoGP architecture.
        EvoGP consists of two main components: Algorithm and Problem. The Algorithm module implements the TGP algorithm along with multiple operation variants. The Problem module integrates various benchmark problems for user evaluation. Within these components, we design CUDA kernels to enable parallel acceleration of computational {processes. These} CUDA kernels are seamlessly integrated into Python using PyTorch's custom operator functionality.
    }
        \label{fig:Structure}
    \end{figure}
}

\newcommand{\figModiNodes}{
    \begin{figure}[b]
        \centering
        \includegraphics[width=0.4\textwidth]
        {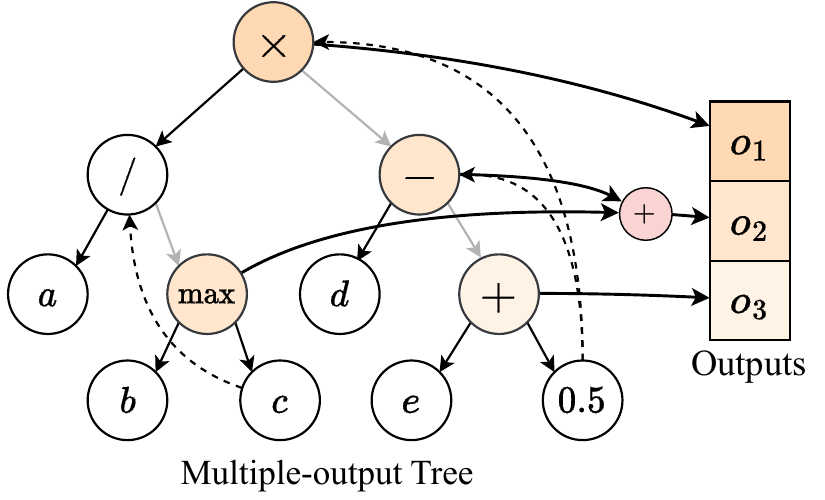}
        \caption{An example of a multiple-output tree. The yellow nodes at different depths represent the Modi nodes. Each Modi node adds its value to the corresponding position in the output and propagates the value of its rightmost subtree to its parent node. The computed process in the figure are: 
        $o_1 = a \div c \times 0.5, o_2 = \max(b, c) + (d - 0.5), o_3 = e + 0.5.$
    }
        \label{fig:modi_nodes}
    \end{figure}
}

\newcommand{\figExperimentTwo}{
    \begin{figure}[t]
        \centering
        \begin{subfigure}{0.485\columnwidth}
            \centering
            \includegraphics[width=\columnwidth]{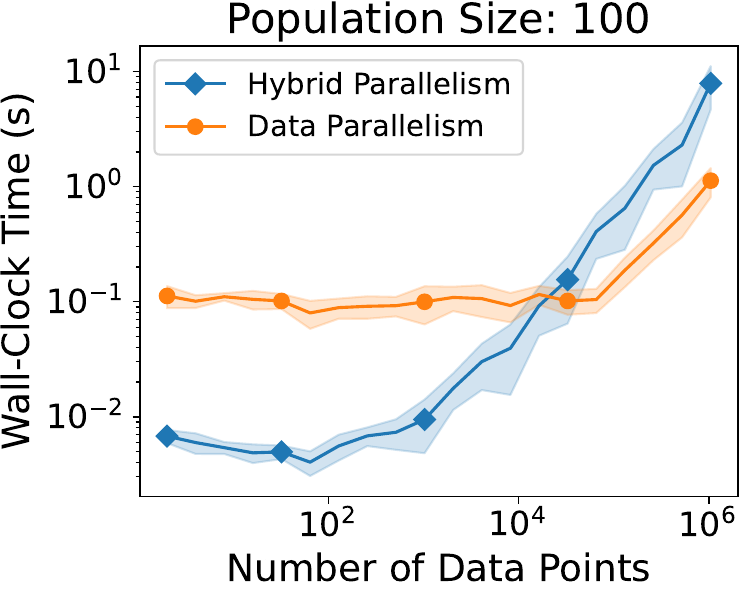}
        \end{subfigure}
        \hfill
        \begin{subfigure}{0.485\columnwidth}
            \centering
            \includegraphics[width=\columnwidth]{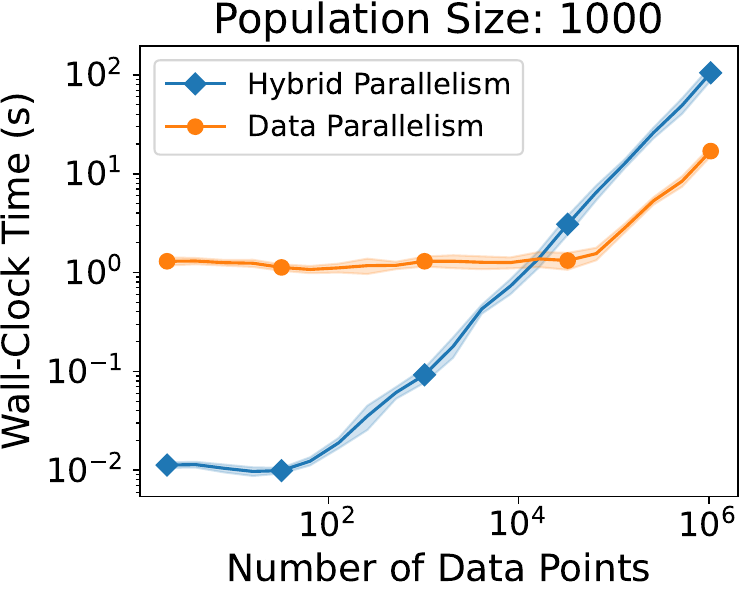}
        \end{subfigure}
        \hfill
        \vspace{0.5em}
        \begin{subfigure}{0.485\columnwidth}
            \centering
            \includegraphics[width=\columnwidth]{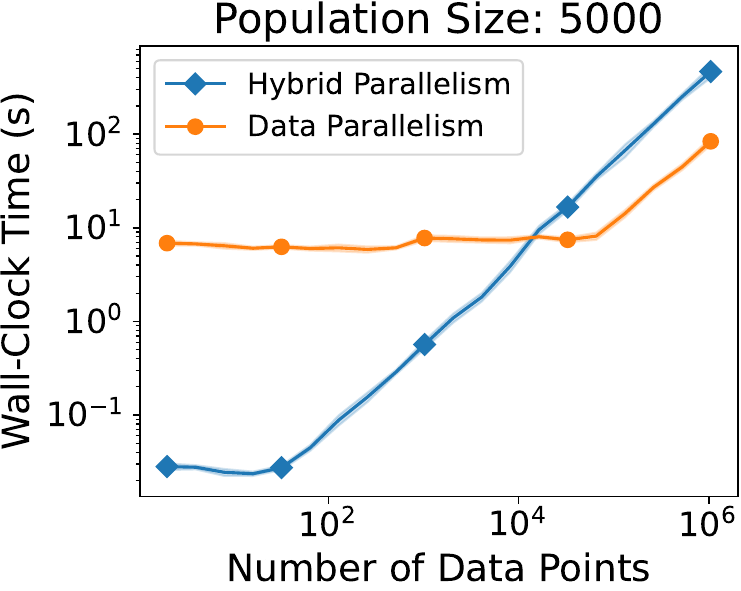}
        \end{subfigure}
        \hfill
        \begin{subfigure}{0.485\columnwidth}
            \centering
            \includegraphics[width=\columnwidth]{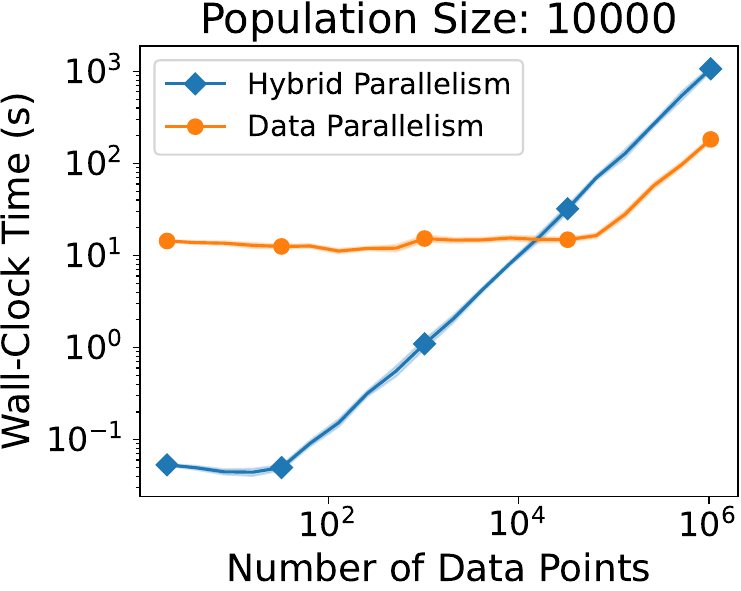}
        \end{subfigure}
        \caption{Parallelism performance in terms of the number of data points.}
        \label{fig:experiment2}
    \end{figure}
}

\newcommand{\figExperimentOneOne}{
    \begin{figure*}[t]
        \centering
        \includegraphics[width=0.94\textwidth]{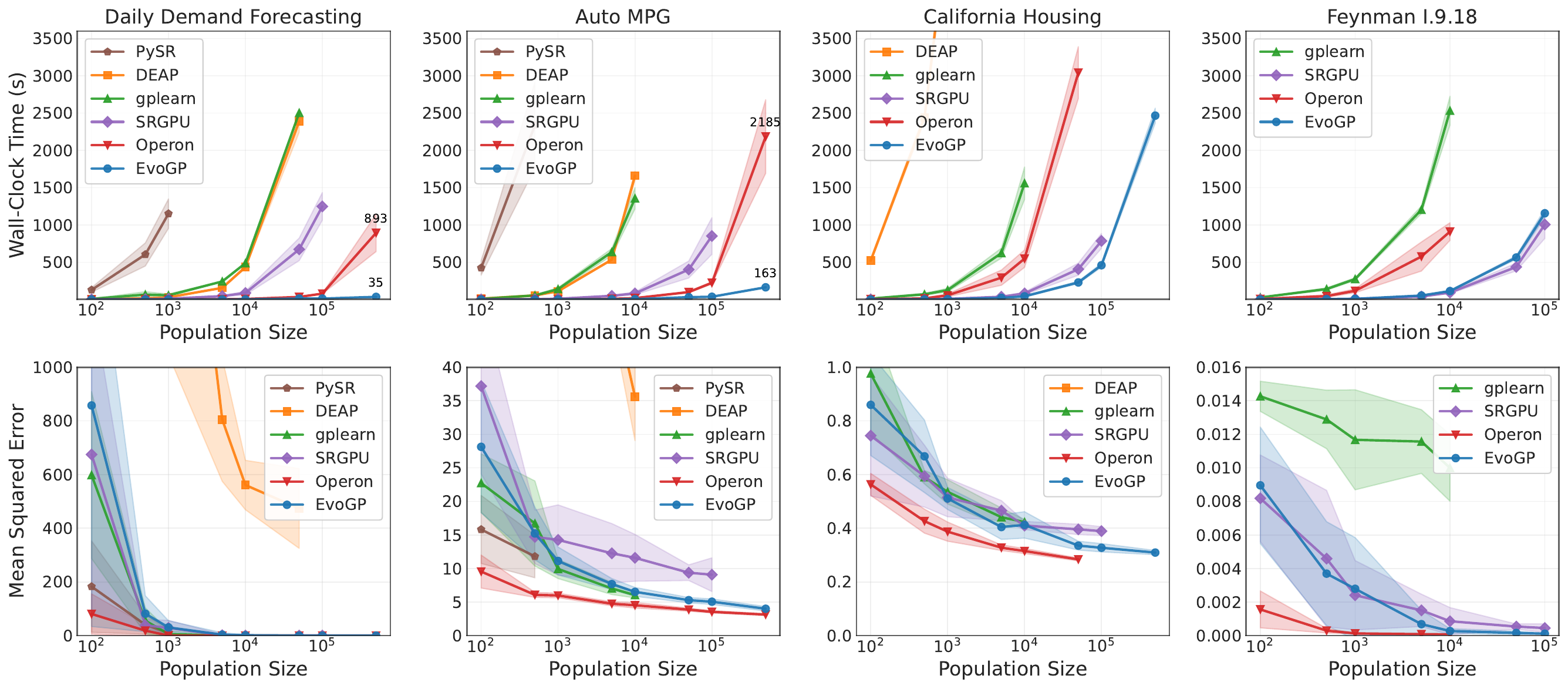}
        \caption{\edit{Comparison with existing TGP implementations on four datasets. The top row presents the total execution time as a function of population size on a logarithmic scale, and the bottom row shows the corresponding solution quality measured by MSE.}}
        \label{fig:experiment1-1}
    \end{figure*}
}

\newcommand{\figExperimentOneTwo}{
    \begin{figure}[t]
        \centering
        \includegraphics[width=0.9\columnwidth]{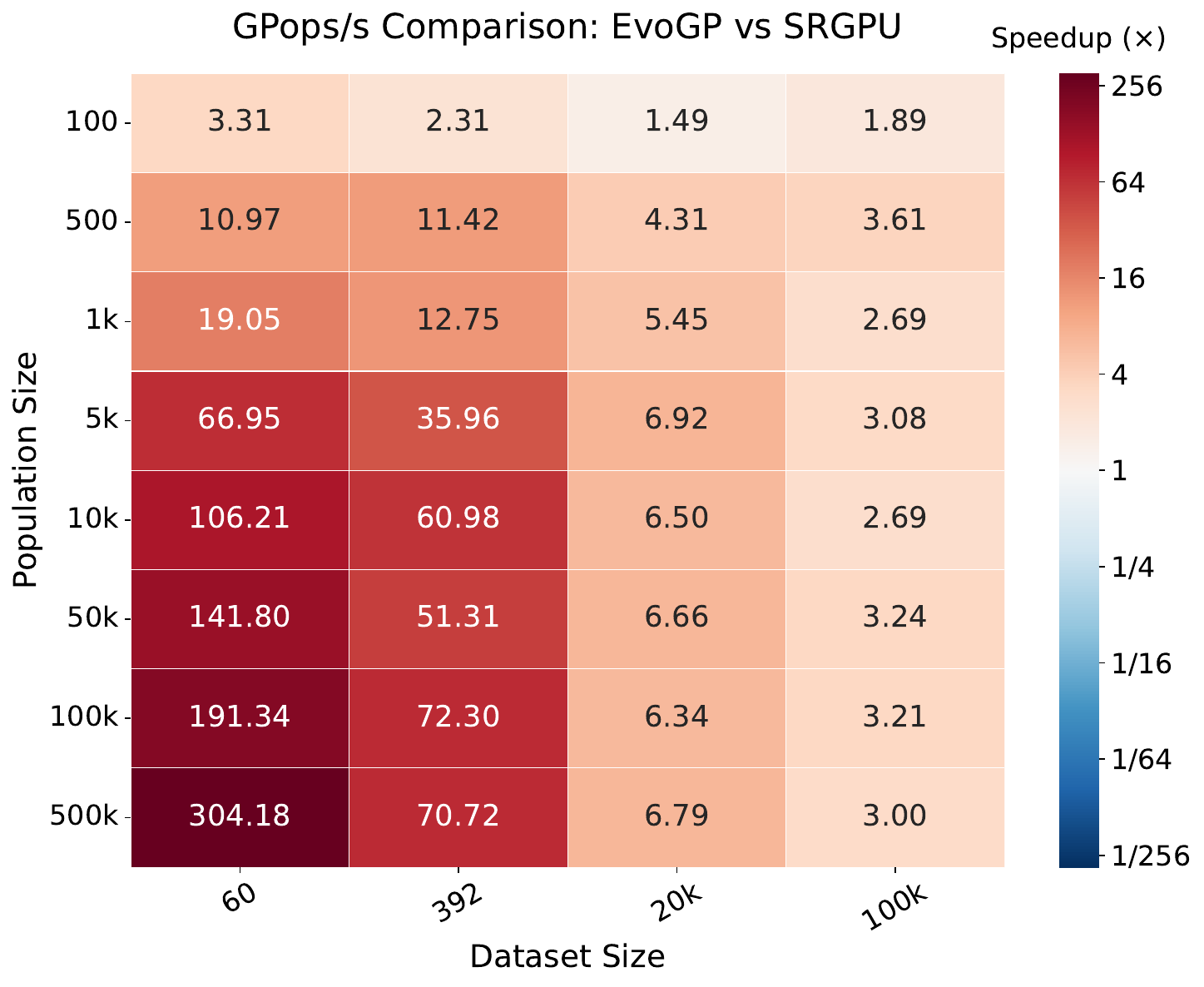}
        \caption{
        {
        Heatmap of GPops/s comparison between EvoGP and SRGPU. 
        Each value represents the speedup factor of EvoGP relative to SRGPU across different population sizes and datasets. 
        Red regions indicate cases where EvoGP achieves higher throughput, whereas blue regions indicate cases where SRGPU performs faster.
        }
        }
        \label{fig:experiment1-2}
    \end{figure}
}

\newcommand{\figExperimentOneThree}{
    \begin{figure}[t]
        \centering
        \includegraphics[width=0.9\columnwidth]{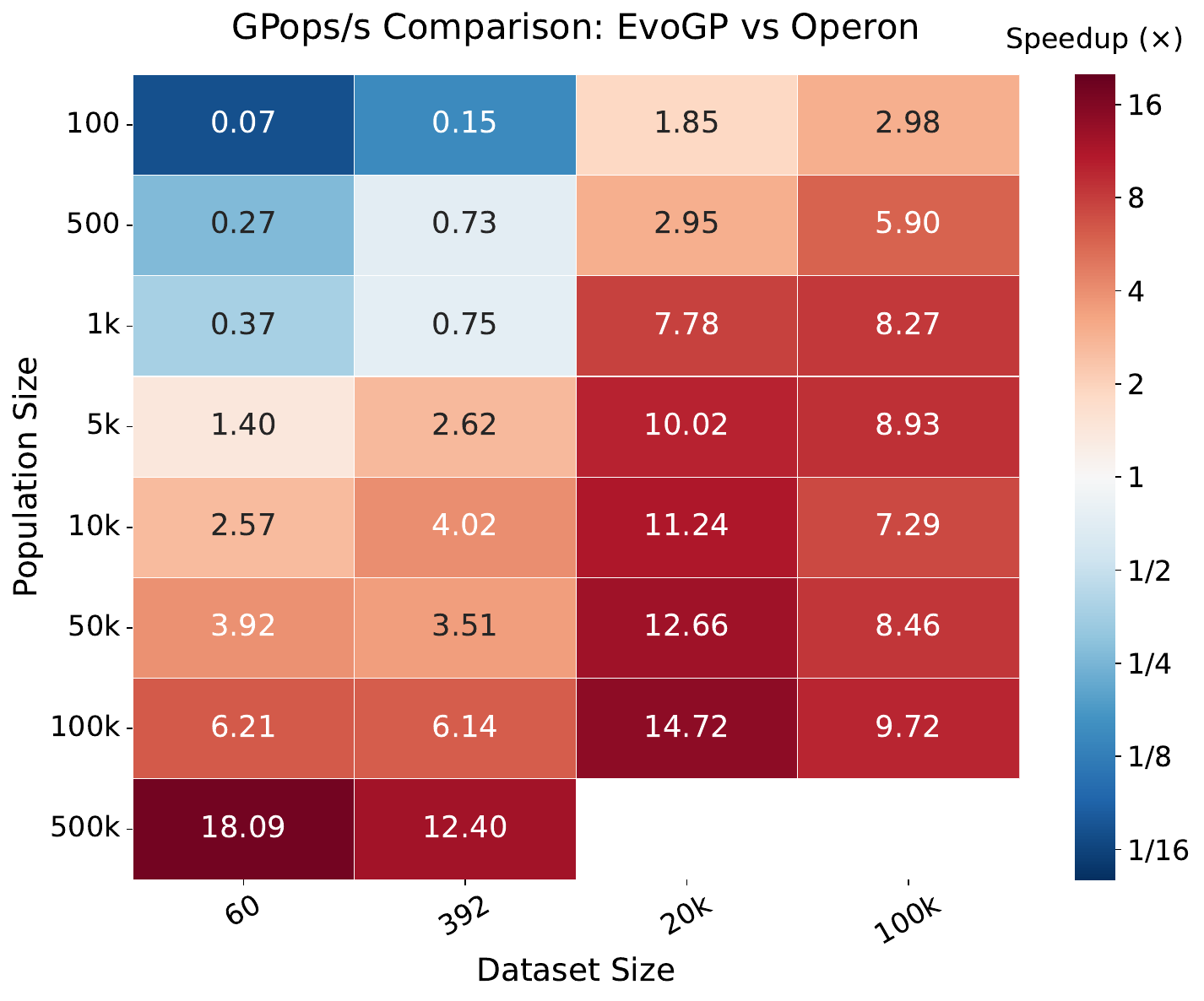}
        \caption{
        {
        Heatmap of GPops/s comparison between EvoGP and Operon. 
        Each value represents the speedup factor of EvoGP relative to Operon across different population sizes and datasets. 
        Red regions indicate cases where EvoGP achieves higher throughput, whereas blue regions indicate cases where Operon performs faster.
        }
        }
        \label{fig:experiment1-3}
    \end{figure}
}

\newcommand{\figExperimentThree}{
    \begin{figure*}[t]
        \centering
        \begin{subfigure}{0.23\textwidth}
            \centering
            \includegraphics[width=\columnwidth]{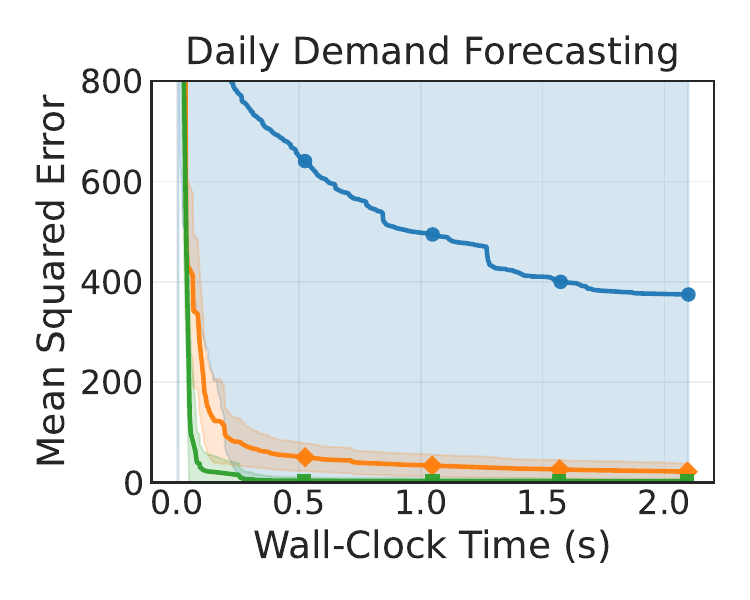}
        \end{subfigure}
        \hfill
        \begin{subfigure}{0.23\textwidth}
            \centering
            \includegraphics[width=\columnwidth]{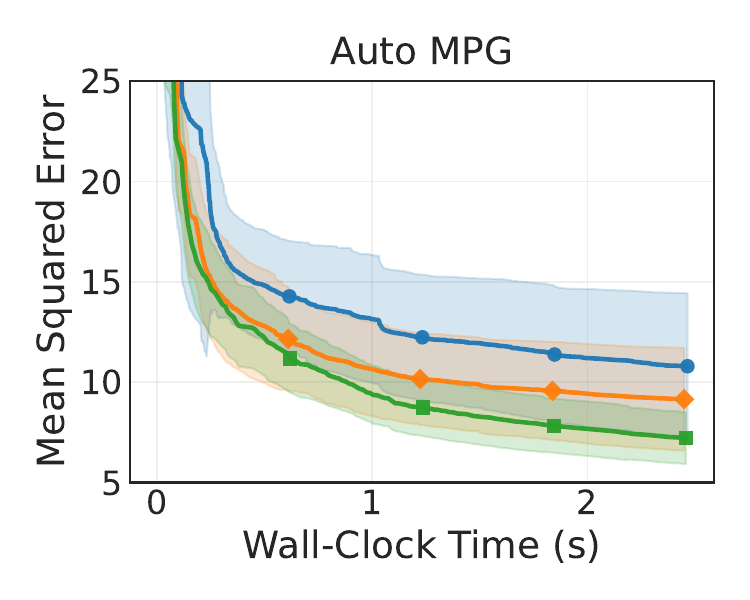}
        \end{subfigure}
        \hfill
        \begin{subfigure}{0.23\textwidth}
            \centering
            \includegraphics[width=\columnwidth]{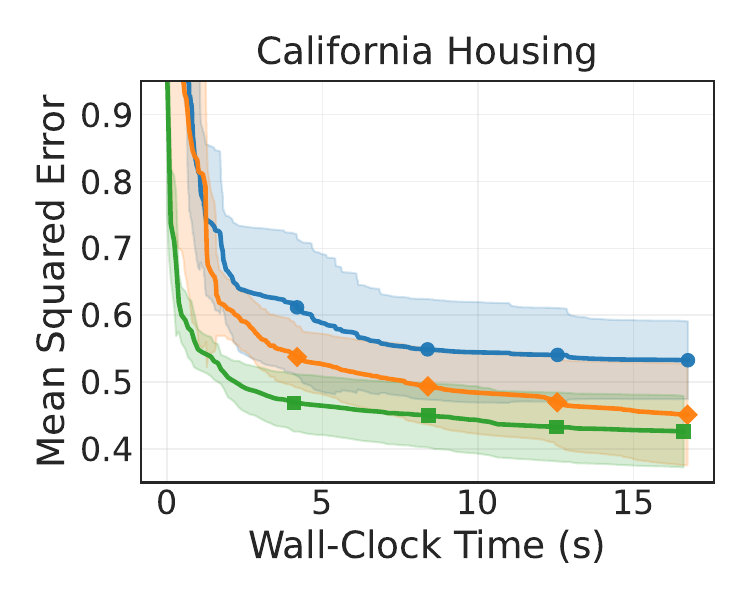}
        \end{subfigure}
        \hfill
        \begin{subfigure}{0.23\textwidth}
            \centering
            \includegraphics[width=\columnwidth]{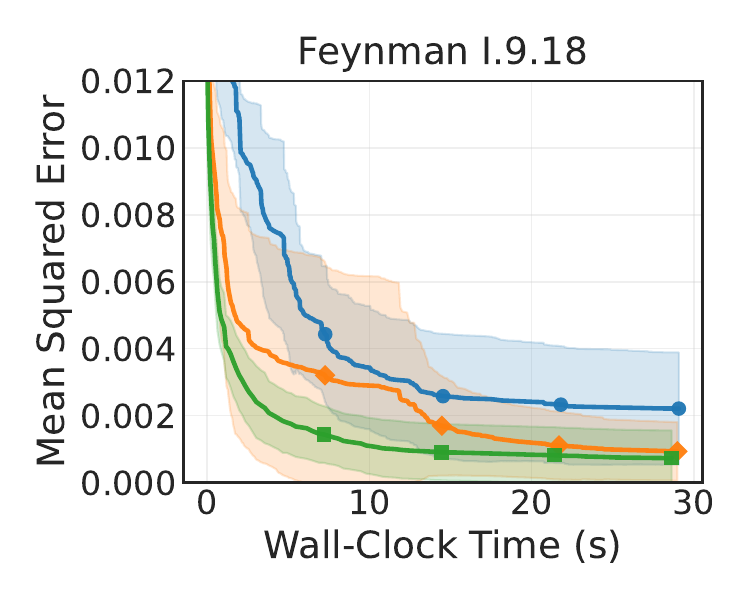}
        \end{subfigure}
        \hfill
        \vspace{0.5em}
        \begin{subfigure}{0.23\textwidth}
            \centering
            \includegraphics[width=\columnwidth]{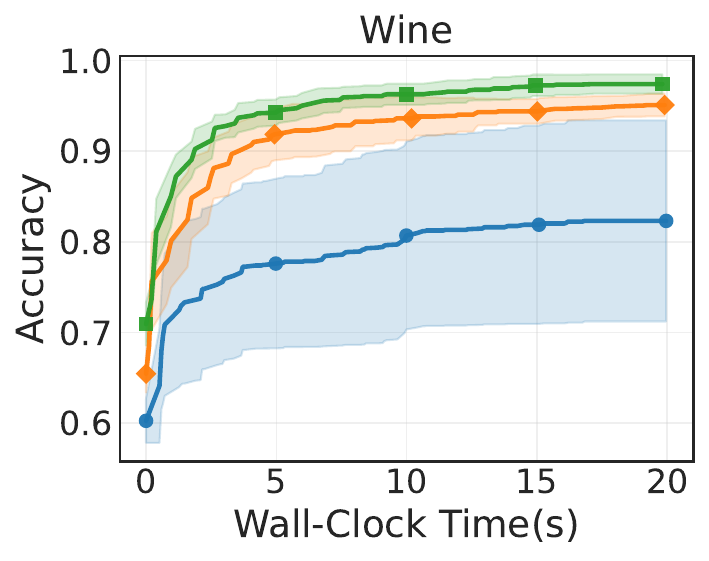}
        \end{subfigure}
        \hfill
        \begin{subfigure}{0.23\textwidth}
            \centering
            \includegraphics[width=\columnwidth]{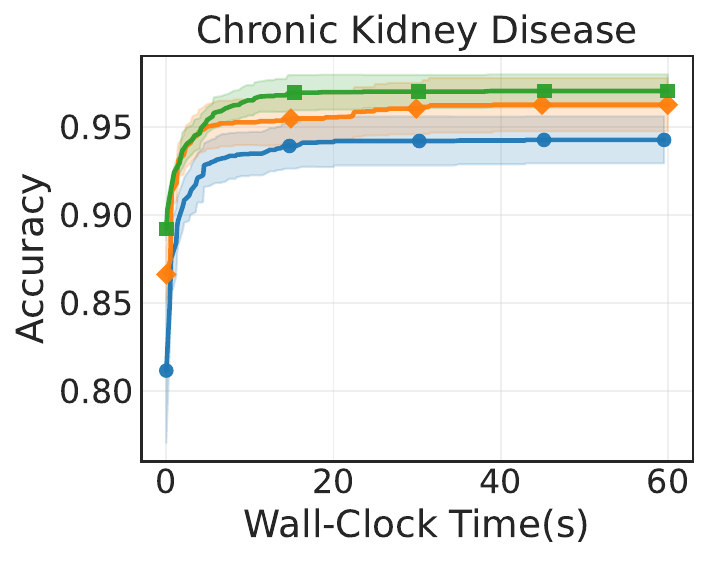}
        \end{subfigure}
        \hfill
        \begin{subfigure}{0.23\textwidth}
            \centering
            \includegraphics[width=\columnwidth]{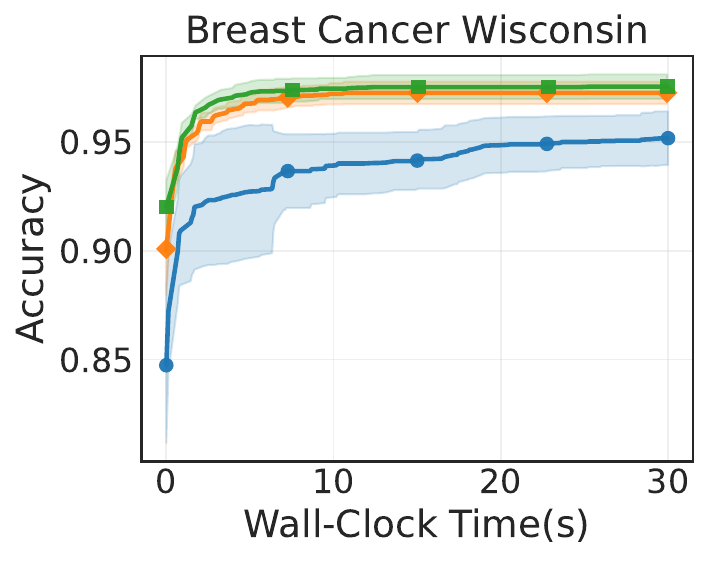}
        \end{subfigure}
        \hfill
        \begin{subfigure}{0.23\textwidth}
            \centering
            \includegraphics[width=\columnwidth]{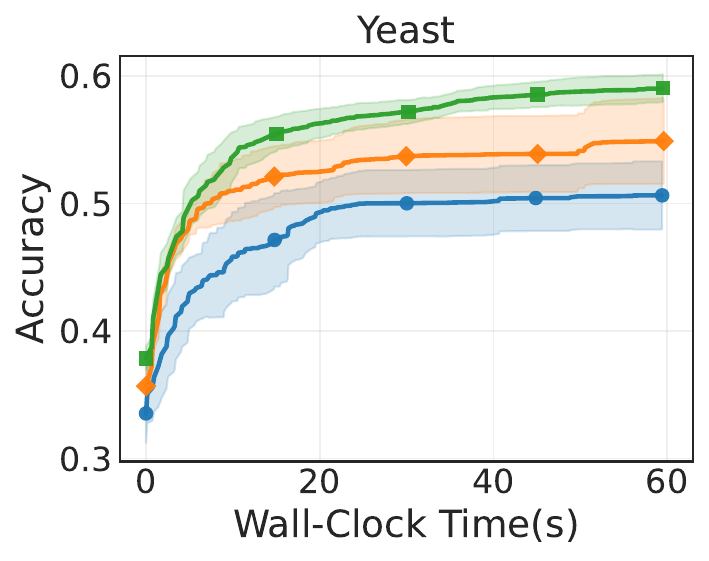}
        \end{subfigure}
        \hfill
        \vspace{0.5em}
        \begin{subfigure}{0.23\textwidth}
            \centering
            \includegraphics[width=\columnwidth]{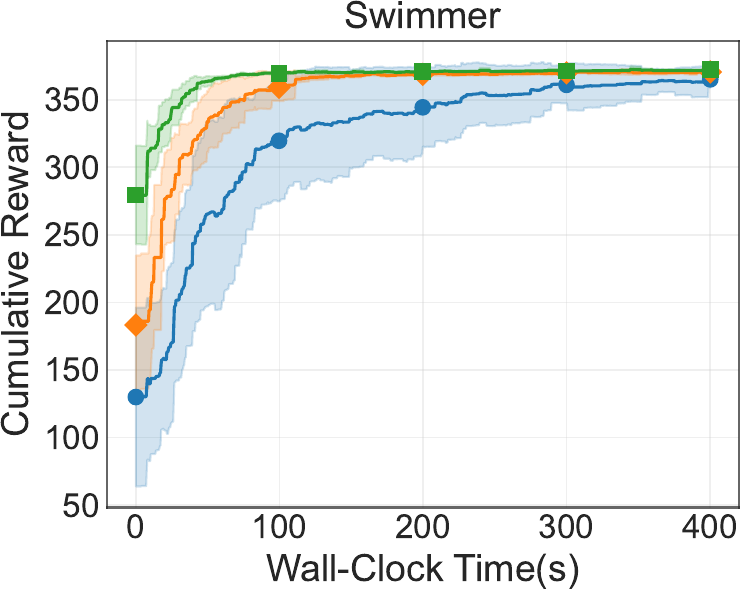}
        \end{subfigure}
        \hfill
        \begin{subfigure}{0.23\textwidth}
            \centering
            \includegraphics[width=\columnwidth]{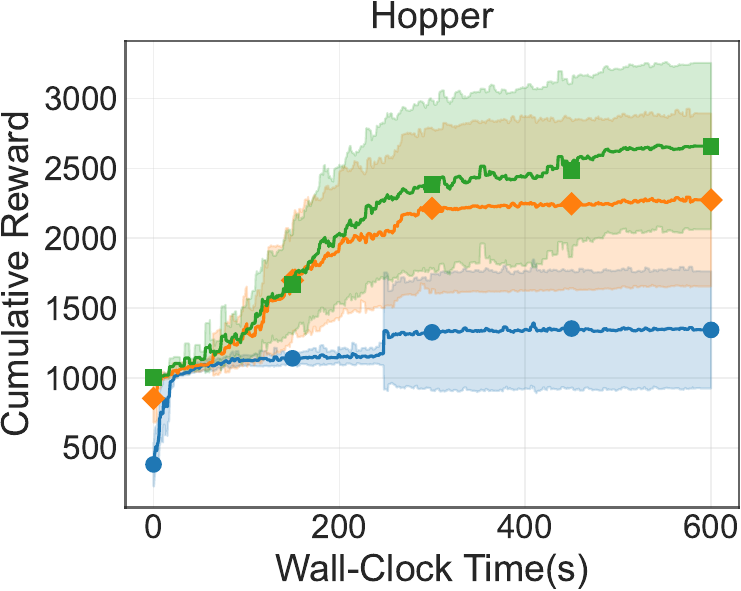}
        \end{subfigure} 
        \hfill
        \begin{subfigure}{0.23\textwidth}
            \centering
            \includegraphics[width=\columnwidth]{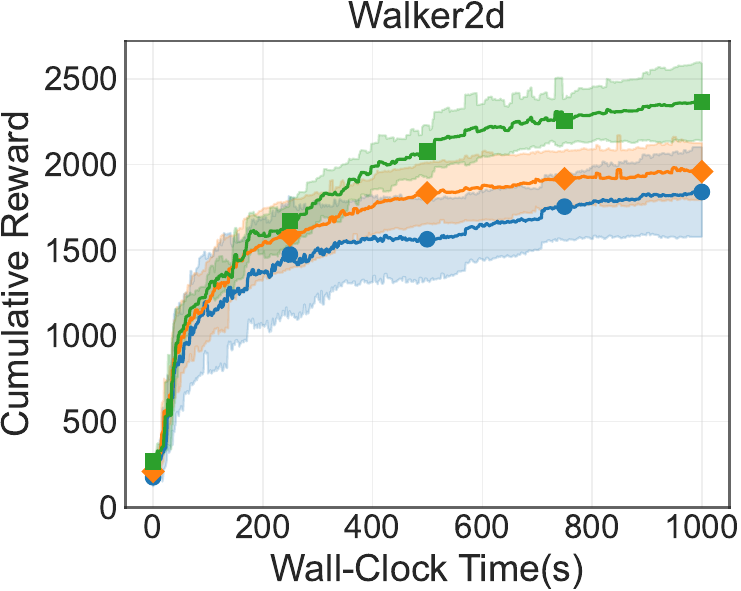}
        \end{subfigure}
        \hfill
        \begin{subfigure}{0.23\textwidth}
            \centering
            \includegraphics[width=\columnwidth]{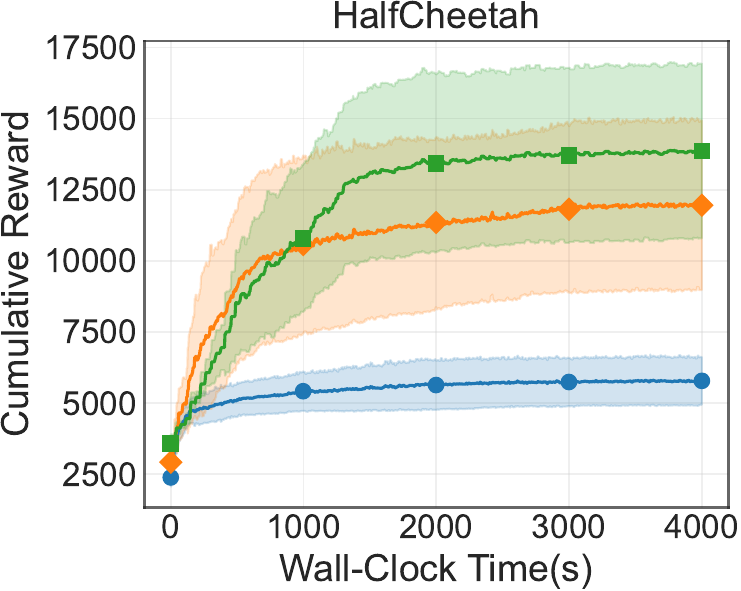}
        \end{subfigure}
        \begin{subfigure}{0.60\textwidth}
            \centering
            \includegraphics[width=\columnwidth]{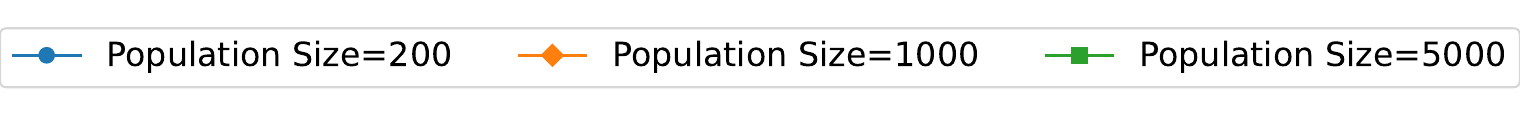}
        \end{subfigure}
        \caption{{Task performance versus wall-clock time for various population sizes. Rows correspond to symbolic regression, classification, and robotics control tasks, respectively.}}
        \label{fig:experiment3}
    \end{figure*}
}

\newcommand{\figTensorgp}{
    \begin{figure*}[hbp]
        \centering
        \begin{subfigure}{0.24\textwidth}
            \centering
            \includegraphics[width=\columnwidth]{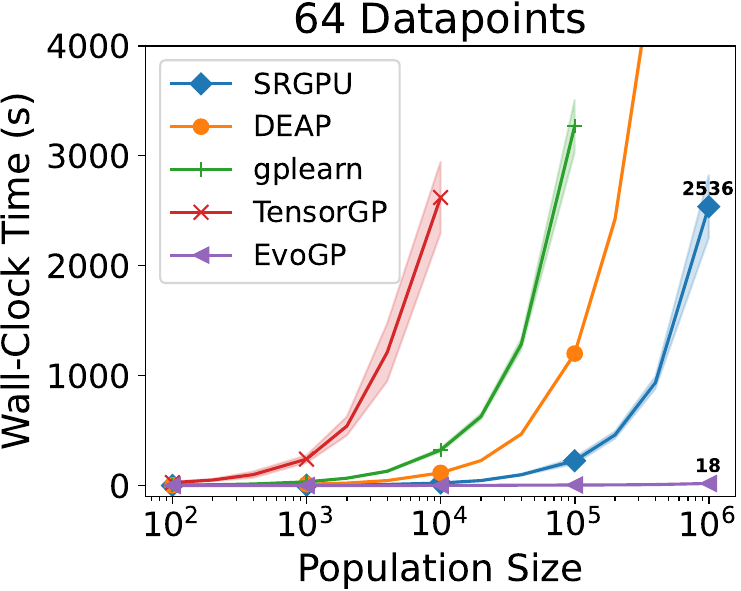}
        \end{subfigure}
        \hfill
        \begin{subfigure}{0.24\textwidth}
            \centering
            \includegraphics[width=\columnwidth]{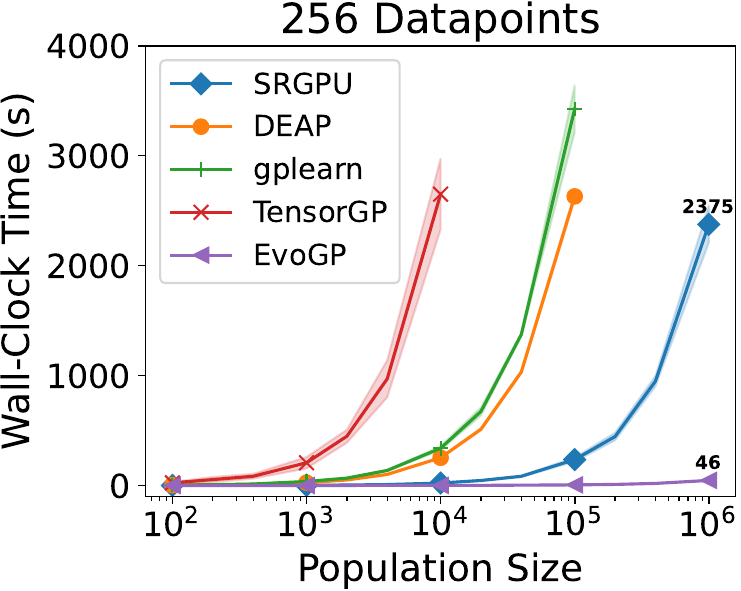}
        \end{subfigure}
        \hfill
        \begin{subfigure}{0.24\textwidth}
            \centering
            \includegraphics[width=\columnwidth]{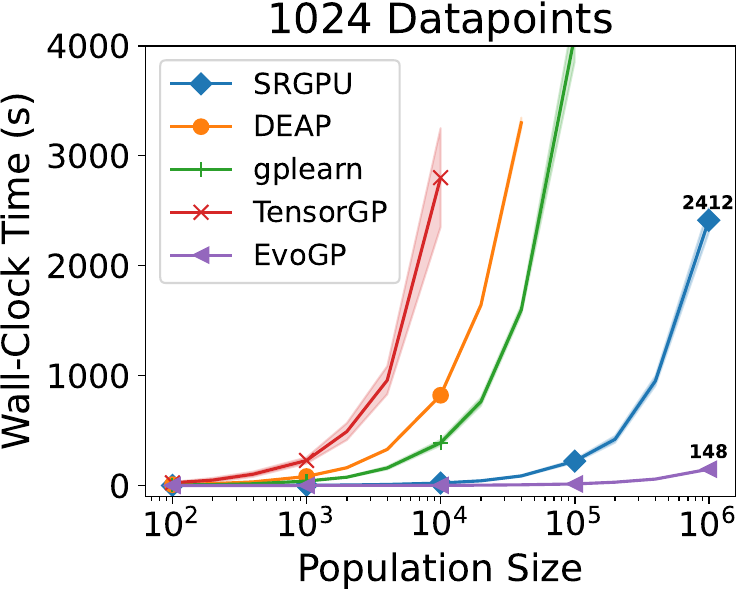}
        \end{subfigure}
        \hfill
        \begin{subfigure}{0.24\textwidth}
            \centering
            \includegraphics[width=\columnwidth]{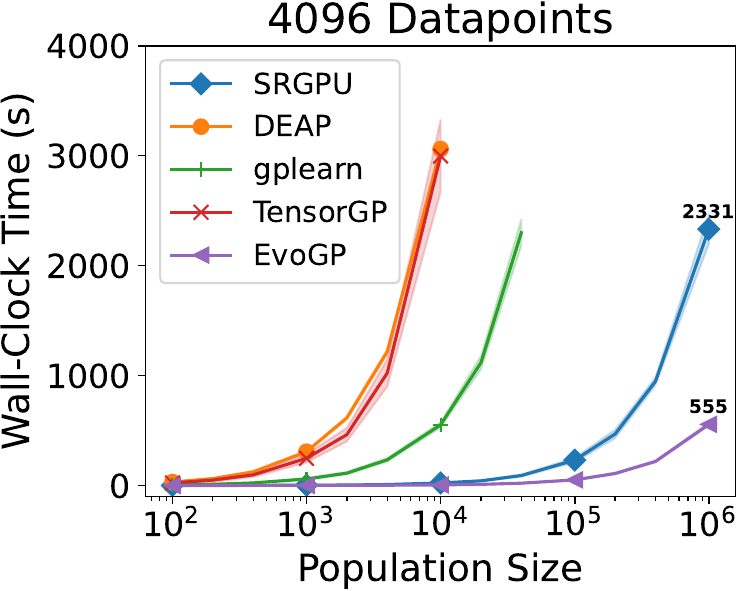}
        \end{subfigure}
        \hfill
        \vspace{0.5em}
        \begin{subfigure}{0.24\textwidth}
            \centering
            \includegraphics[width=\columnwidth]{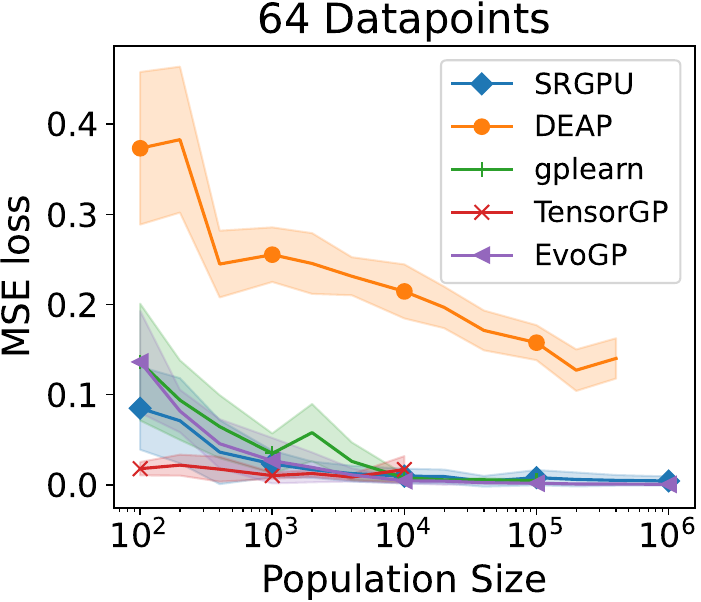}
        \end{subfigure}
        \hfill
        \begin{subfigure}{0.24\textwidth}
            \centering
            \includegraphics[width=\columnwidth]{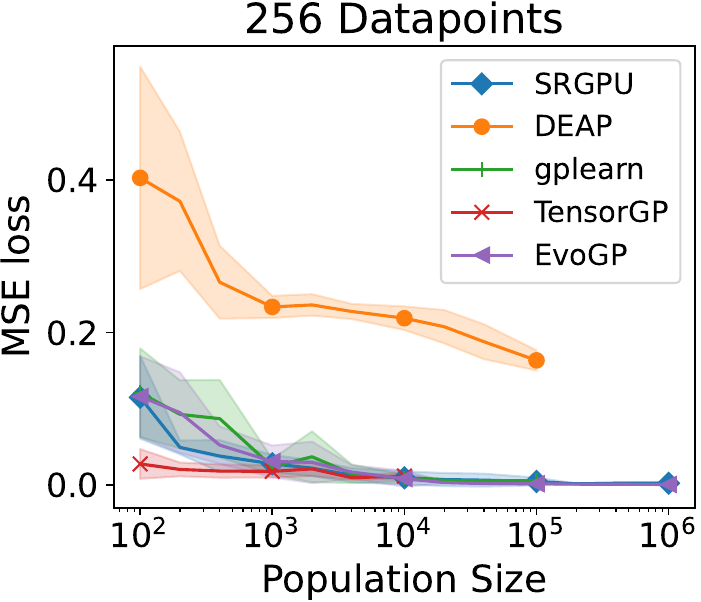}
        \end{subfigure}
        \hfill
        \begin{subfigure}{0.24\textwidth}
            \centering
            \includegraphics[width=\columnwidth]{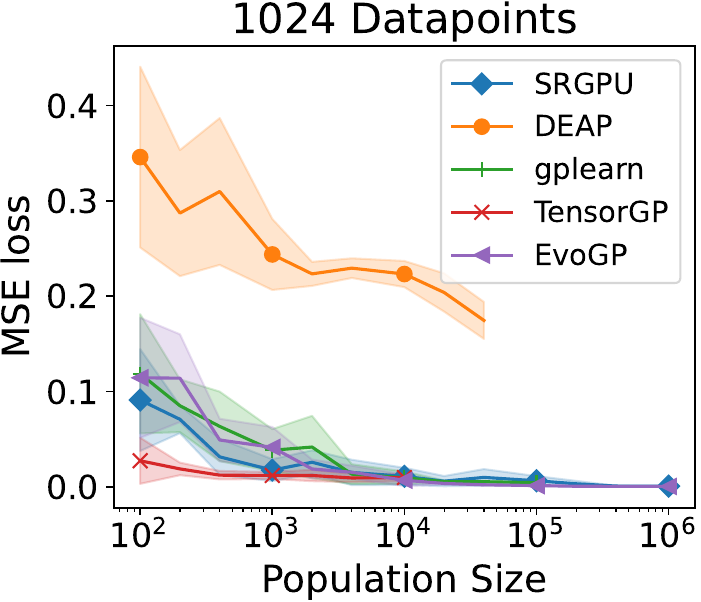}
        \end{subfigure}
        \hfill
        \begin{subfigure}{0.24\textwidth}
            \centering
            \includegraphics[width=\columnwidth]{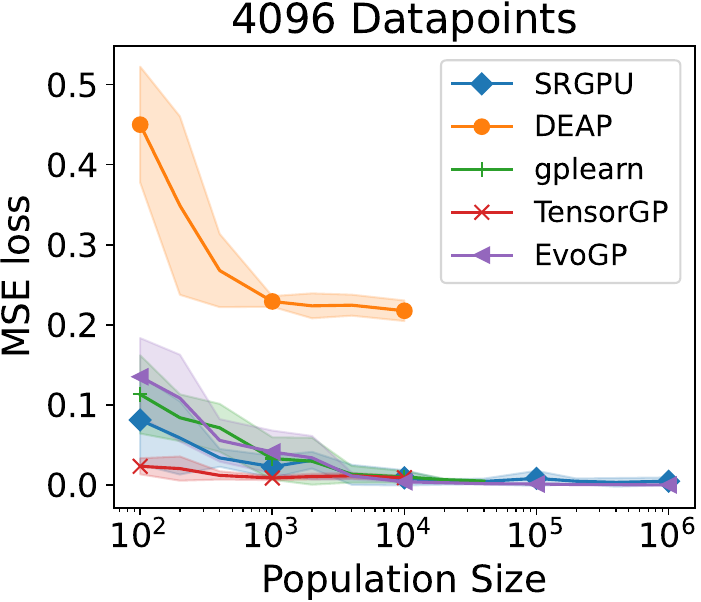}
        \end{subfigure}
            \caption{
        Comparison of EvoGP with TensorGP and other representative libraries on the Pagie polynomial benchmark. 
        The top row shows total execution time as a function of population size, and the bottom row presents solution quality measured by MSE. 
        }
        \label{fig:tensorgp}
    \end{figure*}
}

\newcommand{\algTBGP}{
    \begin{algorithm}[b]
        \caption{Main Process of TGP}
        \label{alg:TBGP}
        \begin{algorithmic}
        \REQUIRE Population size $N$;
                Target fitness $f_{\text{target}}$;
                Max generations count $G$;
        \STATE $P \leftarrow$ Randomly generate $N$ trees;
        \FOR{$g = 1$ to $G$}
            \STATE $Fit \leftarrow$ Evaluate fitness values of $P$;
            \IF{$\max(Fit) \geq f_{\text{target}}$}
            \STATE \textbf{break}
            \ENDIF
            \STATE $C \leftarrow$ The empty population.
            \WHILE{Not enough trees in $C$}
                \STATE $Parent1$, $Parent2$ $\leftarrow$ Select parents;
                \STATE $Child$ $\leftarrow$ Crossover($Parent1$, $Parent2$);
                \STATE $Child$ $\leftarrow$ Mutation($Child$);
                \STATE $C \leftarrow C \cup \{Child\}$
            \ENDWHILE
        \ENDFOR
        \STATE \textbf{return} $P[\arg\max(Fit)]$
        \end{algorithmic}
    \end{algorithm}
}

\newcommand{\tabGpComparison}{
    \begin{table*}[t!]
    \centering
    \caption{Comparison of Mutation and Crossover Operators in Genetic Programming Tools}
    \label{table:gp_comparison}
    \renewcommand{\arraystretch}{1.2}
    \begin{tabular}{ccccccccccc}
    \toprule
    \textbf{Name} & \makecell{\textbf{One Point}\\\textbf{Cross.}} & \makecell{\textbf{One Point}\\\textbf{Leaf Biased}\\\textbf{Cross.}} & \makecell{\textbf{Subtree}\\\textbf{Mut.}} & \makecell{\textbf{Hoist}\\\textbf{Mut.}} & \makecell{\textbf{Single Point}\\\textbf{Mut.}} & \makecell{\textbf{Multi-Point}\\\textbf{Mut.}} & \makecell{\textbf{Insert}\\\textbf{Mut.}} & \makecell{\textbf{Delete}\\\textbf{Mut.}} & \makecell{\textbf{Single Const}\\\textbf{Mut.}} & \makecell{\textbf{Multi-Const}\\\textbf{Mut.}} \\
    \midrule
    DEAP & $\checkmark$ & $\checkmark$ & $\checkmark$ & $\times$ & $\checkmark$ & $\times$ & $\checkmark$ & $\checkmark$ & $\checkmark$ & $\checkmark$ \\
    Gplearn & $\checkmark$ & $\times$ & $\checkmark$ & $\checkmark$ & $\checkmark$ & $\times$ & $\times$ & $\times$ & $\times$ & $\times$ \\
    KarooGP & $\checkmark$ & $\times$ & $\checkmark$ & $\times$ & $\checkmark$ & $\times$ & $\times$ & $\times$ & $\times$ & $\times$ \\
    TensorGP & $\checkmark$ & $\times$ & $\checkmark$ & $\times$ & $\checkmark$ & $\times$ & $\checkmark$ & $\checkmark$ & $\times$ & $\times$ \\
    \makecell{SRGPU} & $\checkmark$ & $\times$ & $\checkmark$ & $\checkmark$ & $\checkmark$ & $\checkmark$ & $\times$ & $\times$ & $\times$ & $\times$ \\
    EvoGP (Ours) & $\checkmark$ & $\checkmark$ & $\checkmark$ & $\checkmark$ & $\checkmark$ & $\checkmark$ & $\checkmark$ & $\checkmark$ & $\checkmark$ & $\checkmark$ \\
    \bottomrule
    \end{tabular}
    \end{table*}
}

\newcommand{\tabSystemSpecifications}{
    \begin{table}[h!]
        \centering
        \caption{{Hardware Specifications}}
        \begin{tabular}{>{\centering}p{4cm} >{\centering\arraybackslash}p{3.5cm}}
            \toprule
            \textbf{Component} & \textbf{Specification} \\
            \midrule
            CPU & Intel(R) Xeon(R) Gold 6226R \\
            CPU Cores & 32 \\
            GPU (Section~\ref{experiment_hybrid}) & NVIDIA RTX 3090 \\
            GPU (Sections~\ref{experiment_comparison} \& \ref{experiment_multitasks}) & NVIDIA RTX 4090\\
            Host RAM & 512 GB \\
            GPU RAM & 24 GB \\
            OS & Ubuntu 22.04.4 LTS \\
            \bottomrule
        \end{tabular}
        \label{tab:system_specifications}
    \end{table}
}

\newcommand{\tabSrParams}{
    \begin{table}[h!]
        \centering
        \caption{{Configuration of Genetic Programming}}
        \begin{tabular}{>{\centering\arraybackslash}p{3.5cm} >{\centering\arraybackslash}p{3.5cm}}
            \toprule
            \textbf{Parameter} & \textbf{Value} \\
            \midrule
            Maximum Tree Size & $512$ \\
            Generations & $100$ \\
            Tournament Size & $20$ \\
            Crossover Probability & $0.9$ \\
            Mutation Probability & $0.1$ \\
            Function Set & \{$+$, $-$, $\times$, $\div$, $\sin$, $\cos$, $\tan$\} \\
            \bottomrule
        \end{tabular}
        \label{tab:sr_params}
    \end{table}
}

\newcommand{\tabSrDatasets}{
    \begin{table}[h!]
        \centering
        \caption{{Details of Symbolic Regression Datasets from the UCI Repository}}
        \begin{tabular}{lcc}
            \toprule
            \textbf{Dataset} & \textbf{Input Features} & \textbf{Data Count} \\
            \midrule
            Daily Demand Forecasting~\cite{daily_demand_forecasting_orders_409}    & 12 & 60   \\
            Auto MPG~\cite{auto_mpg_9}                                             & 7  & 398  \\
            California Housing~\cite{california_housing} & 8  & 20,640 \\
            Feynman I.9.18~\cite{udrescu2020ai}          & 9  & 100,000 \\
            \bottomrule
        \end{tabular}
        \label{tab:sr_datasets}
    \end{table}
}

\newcommand{\tabGpopsSummary}{
    \begin{table}[b]
    \centering
    \caption{Summary of EvoGP GPops/s Performance Across Various Population and Dataset Sizes.}
    \label{tab:gpops_summary}
    \setlength{\tabcolsep}{4pt}
    \begin{tabular}{rcccc}
    \toprule
    \multirow{2}{*}[-2.5pt]{\makecell[r]{\textbf{Population} \\ \textbf{Size}}} & \multicolumn{4}{c}{\textbf{GPops/s by Dataset Size (Datapoints)}} \\
    \cmidrule(lr){2-5}
     & \textbf{60} & \textbf{392} & \textbf{20,640} & \textbf{100,000} \\
    \midrule
    100 & $2.05 \times 10^{8}$ & $1.26 \times 10^{9}$ & $3.49 \times 10^{10}$ & $1.55 \times 10^{11}$ \\
    500 & $1.20 \times 10^{9}$ & $8.67 \times 10^{9}$ & $1.27 \times 10^{11}$ & $3.35 \times 10^{11}$ \\
    1,000 & $1.85 \times 10^{9}$ & $1.02 \times 10^{10}$ & $1.61 \times 10^{11}$ & $3.50 \times 10^{11}$ \\
    5,000 & $7.31 \times 10^{9}$ & $2.93 \times 10^{10}$ & $2.07 \times 10^{11}$ & $4.14 \times 10^{11}$ \\
    10,000 & $1.13 \times 10^{10}$ & $4.32 \times 10^{10}$ & $2.08 \times 10^{11}$ & $3.92 \times 10^{11}$ \\
    50,000 & $1.67 \times 10^{10}$ & $3.38 \times 10^{10}$ & $2.14 \times 10^{11}$ & $4.10 \times 10^{11}$ \\
    100,000 & $2.27 \times 10^{10}$ & $5.31 \times 10^{10}$ & $2.12 \times 10^{11}$ & $4.05 \times 10^{11}$ \\
    500,000 & $3.25 \times 10^{10}$ & $5.87 \times 10^{10}$ & $1.99 \times 10^{11}$ & $3.78 \times 10^{11}$ \\
    1,000,000 & $3.29 \times 10^{10}$ & $5.87 \times 10^{10}$ & $1.97 \times 10^{11}$ & $3.90 \times 10^{11}$ \\
    \bottomrule
    \end{tabular}
    \end{table}
}

\newcommand{\tabClassificationDatasets}{
    \begin{table}[t]
        \centering
        \caption{{Details of Classification Datasets}}
        \label{tab:classification_datasets}
        \setlength{\tabcolsep}{4pt}
        \begin{tabular}{
            >{\raggedright\arraybackslash}m{4cm}
            >{\centering\arraybackslash}m{1.2cm}
            >{\centering\arraybackslash}m{1.2cm}
            >{\centering\arraybackslash}m{1.2cm}
        }
            \toprule
            \textbf{Dataset} & \textbf{Input Features} & \textbf{Output Classes} & \textbf{Data Count} \\
            \midrule
            Wine~\cite{wine_109} & 13 & 3 & 178 \\
            Chronic Kidney Disease~\cite{chronic_kidney_disease_336} & 24 & 3 & 400 \\
            Breast Cancer Wisconsin~\cite{breast_cancer_wisconsin_(diagnostic)_17} & 30 & 2 & 569 \\
            Yeast~\cite{yeast_110} & 8 & 10 & 1484 \\
            \bottomrule
        \end{tabular}
    \end{table}
}

\newcommand{\tabBraxEnvironmentDimensions}{
    \begin{table}[t]
        \centering
        \caption{\centering {Details of Brax Environments}}
        \begin{tabular}{lcc}
            \toprule
            \textbf{Environment} & \textbf{Observation Dimension} & \textbf{Action Dimension} \\
            \midrule
            Swimmer      & 8  & 2  \\
            Hopper       & 17 & 3  \\
            Walker2d     & 17 & 6  \\
            HalfCheetah  & 17 & 6  \\
            \bottomrule
        \end{tabular}
        \label{tab:brax_environment_dimensions}
    \end{table}
}

\newcommand{\tabPerformanceSummaryWide}{
    \begin{table}[b]
    \centering
    \begin{threeparttable}
    \caption{{Performance Across Various Tasks and Population Sizes in the Same Time.}}
    \label{tab:performance_summary_wide}
    \setlength{\tabcolsep}{6pt}
    
    \begin{tabular}{l c c c}
        \toprule
        \multirow{2}{*}{\textbf{Task Identifier}} & \multicolumn{3}{c}{\textbf{Population Size}} \\
        \cmidrule(lr){2-4}
        & \textbf{200} & \textbf{1,000} & \textbf{5,000} \\
        \midrule
        \multicolumn{4}{l}{\textit{Symbolic Regression (MSE Loss --- lower is better)}} \\
        \quad Daily Demand Forecasting & 374.578 & 21.330 & \textbf{1.963} \\
        \quad Auto MPG & 10.790 & 9.144 & \textbf{7.205} \\
        \quad California Housing & 0.533 & 0.451 & \textbf{0.426} \\
        \quad Feynman I.9.18 & 2.207e-3 & 9.274e-4 & \textbf{7.239e-4} \\
        \midrule
        \multicolumn{4}{l}{\textit{Classification (Accuracy --- higher is better)}} \\
        \quad Wine & 0.823 & 0.951 & \textbf{0.974} \\
        \quad Chronic Kidney Disease & 0.943 & 0.963 & \textbf{0.971} \\
        \quad Breast Cancer Wisconsin & 0.952 & 0.973 & \textbf{0.976} \\
        \quad Yeast & 0.506 & 0.549 & \textbf{0.591} \\
        \midrule
        \multicolumn{4}{l}{\textit{Robotics Control (Cumulative Reward --- higher is better)}} \\
        \quad Swimmer & 365.0 & 370.5 & \textbf{372.2} \\
        \quad Hopper & 1344.7 & 2274.4 & \textbf{2658.9} \\
        \quad Walker2d & 1839.5 & 1958.5 & \textbf{2364.7} \\
        \quad Halfcheetah & 5781.2 & 11954.0 & \textbf{13862.9} \\
        \bottomrule
    \end{tabular}
    
    \begin{tablenotes}
        \item[*] The best performances are highlighted in bold.
    \end{tablenotes}
    
    \end{threeparttable}
    \end{table}
}

\newcommand{\tabPySRBreakdown}{
    \begin{table}[h]
    \centering
    \setlength{\tabcolsep}{2.5pt}
    \renewcommand{\arraystretch}{1.3}
    \caption{Detailed Breakdown of PySR GPops/s Performance Across Population and Dataset Sizes}
    \begin{tabular}{|r|c c c|c c c|c c c|c c c|}
    \hline
    \multirow{2}{*}[-1pt]{\makecell[r]{\textbf{Population} \\ \textbf{Size}}} & \multicolumn{3}{c|}{\textbf{60 Datapoints}} & \multicolumn{3}{c|}{\textbf{392 Datapoints}} & \multicolumn{3}{c|}{\textbf{20640 Datapoints}} & \multicolumn{3}{c|}{\textbf{100000 Datapoints}} \\
    \cline{2-13}
    & \textbf{Time} & \textbf{Avg. Size} & \textbf{GPops/s} & \textbf{Time} & \textbf{Avg. Size} & \textbf{GPops/s} & \textbf{Time} & \textbf{Avg. Size} & \textbf{GPops/s} & \textbf{Time} & \textbf{Avg. Size} & \textbf{GPops/s} \\
    \hline
    100 & 127.94 & 333.71 & $1.68 \times 10^{6}$ & 423.20 & 381.91 & $3.56 \times 10^{6}$ & 3650.04 & 392.05 & $2.22 \times 10^{7}$ & 4164.54 & 463.53 & $1.11 \times 10^{8}$ \\
    500 & 603.14 & 288.50 & $1.36 \times 10^{6}$ & 2337.47 & 229.55 & $2.21 \times 10^{6}$ & -- & -- & -- & -- & -- & -- \\
    1,000 & 1150.52 & 334.77 & $1.87 \times 10^{6}$ & -- & -- & -- & -- & -- & -- & -- & -- & -- \\
    \hline
    \end{tabular}
    \end{table}
}

\newcommand{\tabgplearnBreakdown}{
    \begin{table}[h]
        \centering
        \setlength{\tabcolsep}{2.5pt}
        \renewcommand{\arraystretch}{1.3}
        \caption{Detailed Breakdown of gplearn GPops/s Performance Across Population and Dataset Sizes}
        \begin{tabular}{|r|c c c|c c c|c c c|c c c|}
            \hline
            \multirow{2}{*}[-1pt]{\makecell[r]{\textbf{Population} \\ \textbf{Size}}} & 
            \multicolumn{3}{c|}{\textbf{60 Datapoints}} & 
            \multicolumn{3}{c|}{\textbf{392 Datapoints}} & 
            \multicolumn{3}{c|}{\textbf{20640 Datapoints}} & 
            \multicolumn{3}{c|}{\textbf{100000 Datapoints}} \\
            \cline{2-13}
            & \textbf{Time} & \textbf{Avg. Size} & \textbf{GPops/s} 
            & \textbf{Time} & \textbf{Avg. Size} & \textbf{GPops/s} 
            & \textbf{Time} & \textbf{Avg. Size} & \textbf{GPops/s} 
            & \textbf{Time} & \textbf{Avg. Size} & \textbf{GPops/s} \\
            \hline
            100    & 7.95    & 421.27  & $2.16 \times 10^{7}$ & 7.73    & 213.04 & $9.63 \times 10^{7}$ & 11.62   & 16.69  & $2.46 \times 10^{8}$ & 26.56   & 4.14 & $1.58 \times 10^{8}$ \\
            500    & 65.16   & 1123.00 & $3.01 \times 10^{7}$ & 53.74   & 334.33 & $1.13 \times 10^{8}$ & 69.02   & 15.02  & $2.09 \times 10^{8}$ & 139.54  & 3.67 & $1.40 \times 10^{8}$ \\
            1,000  & 55.89   & 109.76  & $5.56 \times 10^{6}$ & 138.88  & 446.65 & $1.22 \times 10^{8}$ & 125.72  & 17.51  & $2.74 \times 10^{8}$ & 273.49  & 3.90 & $1.46 \times 10^{8}$ \\
            5,000  & 241.29  & 5.28    & $6.57 \times 10^{5}$ & 635.97  & 393.31 & $1.21 \times 10^{8}$ & 621.58  & 24.95  & $4.21 \times 10^{8}$ & 1203.28 & 3.97 & $1.66 \times 10^{8}$ \\
            10,000 & 486.40  & 5.27    & $6.50 \times 10^{5}$ & 1356.35 & 440.97 & $1.27 \times 10^{8}$ & 1558.67 & 27.55  & $3.75 \times 10^{8}$ & 2531.53 & 4.71 & $1.82 \times 10^{8}$ \\
            50,000 & 2502.93 & 5.25    & $6.30 \times 10^{5}$ & --      & --     & --                   & --      & --     & --                   & --      & --   & --                   \\
            \hline
        \end{tabular}
    \end{table}
}

\newcommand{\tabDEAPBreakdown}{
    \begin{table}[h]
        \centering
        \setlength{\tabcolsep}{2.5pt}
        \renewcommand{\arraystretch}{1.3}
        \caption{Detailed Breakdown of DEAP GPops/s Performance Across Population and Dataset Sizes}
        \begin{tabular}{|r|c c c|c c c|c c c|c c c|}
            \hline
            \multirow{2}{*}[-1pt]{\makecell[r]{\textbf{Population} \\ \textbf{Size}}} & 
            \multicolumn{3}{c|}{\textbf{60 Datapoints}} & 
            \multicolumn{3}{c|}{\textbf{392 Datapoints}} & 
            \multicolumn{3}{c|}{\textbf{20640 Datapoints}} & 
            \multicolumn{3}{c|}{\textbf{100000 Datapoints}} \\
            \cline{2-13}
            & \textbf{Time} & \textbf{Avg. Size} & \textbf{GPops/s} 
            & \textbf{Time} & \textbf{Avg. Size} & \textbf{GPops/s} 
            & \textbf{Time} & \textbf{Avg. Size} & \textbf{GPops/s} 
            & \textbf{Time} & \textbf{Avg. Size} & \textbf{GPops/s} \\
            \hline
            100    & 3.10    & 9.15 & $1.78 \times 10^{6}$ & 11.01   & 9.02 & $3.22 \times 10^{6}$ & 521.42  & 9.90 & $3.90 \times 10^{6}$ & 2304.46 & 8.48 & $3.68 \times 10^{6}$ \\
            500    & 14.46   & 9.02 & $1.87 \times 10^{6}$ & 53.78   & 9.16 & $3.34 \times 10^{6}$ & 2439.16 & 9.24 & $3.91 \times 10^{6}$ & --      & --   & --                   \\
            1,000  & 29.03   & 9.36 & $1.94 \times 10^{6}$ & 109.25  & 9.34 & $3.35 \times 10^{6}$ & 4978.83 & 9.29 & $3.86 \times 10^{6}$ & --      & --   & --                   \\
            5,000  & 155.35  & 9.33 & $1.80 \times 10^{6}$ & 534.29  & 9.32 & $3.42 \times 10^{6}$ & --      & --   & --                   & --      & --   & --                   \\
            10,000 & 433.78  & 9.26 & $1.28 \times 10^{6}$ & 1659.54 & 9.36 & $2.22 \times 10^{6}$ & --      & --   & --                   & --      & --   & --                   \\
            50,000 & 2388.42 & 9.29 & $1.17 \times 10^{6}$ & --      & --   & --                   & --      & --   & --                   & --      & --   & --                   \\
            \hline
        \end{tabular}
    \end{table}
}

\newcommand{\tabSRGPUBreakdown}{
    \begin{table}[h]
        \centering
        \setlength{\tabcolsep}{2.5pt}
        \renewcommand{\arraystretch}{1.3}
        \caption{Detailed Breakdown of SRGPU GPops/s Performance Across Population and Dataset Sizes}
        \begin{tabular}{|r|c c c|c c c|c c c|c c c|}
            \hline
            \multirow{2}{*}[-1pt]{\makecell[r]{\textbf{Population} \\ \textbf{Size}}} & 
            \multicolumn{3}{c|}{\textbf{60 Datapoints}} & 
            \multicolumn{3}{c|}{\textbf{392 Datapoints}} & 
            \multicolumn{3}{c|}{\textbf{20640 Datapoints}} & 
            \multicolumn{3}{c|}{\textbf{100000 Datapoints}} \\
            \cline{2-13}
            & \textbf{Time} & \textbf{Avg. Size} & \textbf{GPops/s} 
            & \textbf{Time} & \textbf{Avg. Size} & \textbf{GPops/s} 
            & \textbf{Time} & \textbf{Avg. Size} & \textbf{GPops/s} 
            & \textbf{Time} & \textbf{Avg. Size} & \textbf{GPops/s} \\
            \hline
            100     & 0.63    & 76.50  & $6.17 \times 10^{7}$ & 0.58    & 93.43  & $5.48 \times 10^{8}$ & 1.07    & 134.39 & $2.35 \times 10^{10}$ & 0.70    & 60.97  & $8.21 \times 10^{10}$ \\
            500     & 4.15    & 150.38 & $1.10 \times 10^{8}$ & 3.88    & 151.82 & $7.59 \times 10^{8}$ & 4.06    & 116.70 & $2.94 \times 10^{10}$ & 3.71    & 81.91  & $9.30 \times 10^{10}$ \\
            1,000   & 9.98    & 193.33 & $9.70 \times 10^{7}$ & 9.56    & 202.40 & $7.99 \times 10^{8}$ & 7.59    & 111.87 & $2.95 \times 10^{10}$ & 8.80    & 123.32 & $1.30 \times 10^{11}$ \\
            5,000   & 44.69   & 164.50 & $1.09 \times 10^{8}$ & 48.28   & 202.83 & $8.14 \times 10^{8}$ & 36.25   & 106.71 & $3.00 \times 10^{10}$ & 38.79   & 105.19 & $1.34 \times 10^{11}$ \\
            10,000  & 86.18   & 166.69 & $1.07 \times 10^{8}$ & 81.19   & 153.79 & $7.08 \times 10^{8}$ & 80.98   & 127.64 & $3.20 \times 10^{10}$ & 92.21   & 133.77 & $1.46 \times 10^{11}$ \\
            50,000  & 671.76  & 265.00 & $1.18 \times 10^{8}$ & 401.45  & 141.05 & $6.59 \times 10^{8}$ & 406.17  & 128.69 & $3.21 \times 10^{10}$ & 435.76  & 111.55 & $1.26 \times 10^{11}$ \\
            100,000 & 1247.69 & 249.99 & $1.19 \times 10^{8}$ & 851.06  & 167.16 & $7.35 \times 10^{8}$ & 784.60  & 128.64 & $3.35 \times 10^{10}$ & 1004.88 & 129.30 & $1.26 \times 10^{11}$ \\
            500,000 & 6079.53 & 218.63 & $1.07 \times 10^{8}$ & 5248.66 & 227.63 & $8.30 \times 10^{8}$ & 4088.10 & 116.60 & $2.93 \times 10^{10}$ & 4677.27 & 118.71 & $1.26 \times 10^{11}$ \\
            \hline
        \end{tabular}
    \end{table}
}

\newcommand{\tabOperonBreakdown}{
    \begin{table}[h]
        \centering
        \setlength{\tabcolsep}{2.5pt}
        \renewcommand{\arraystretch}{1.3}
        \caption{Detailed Breakdown of Operon GPops/s Performance Across Population and Dataset Sizes}
        \begin{tabular}{|r|c c c|c c c|c c c|c c c|}
            \hline
            \multirow{2}{*}[-1pt]{\makecell[r]{\textbf{Population} \\ \textbf{Size}}} & 
            \multicolumn{3}{c|}{\textbf{60 Datapoints}} & 
            \multicolumn{3}{c|}{\textbf{392 Datapoints}} & 
            \multicolumn{3}{c|}{\textbf{20640 Datapoints}} & 
            \multicolumn{3}{c|}{\textbf{100000 Datapoints}} \\
            \cline{2-13}
            & \textbf{Time} & \textbf{Avg. Size} & \textbf{GPops/s} 
            & \textbf{Time} & \textbf{Avg. Size} & \textbf{GPops/s} 
            & \textbf{Time} & \textbf{Avg. Size} & \textbf{GPops/s} 
            & \textbf{Time} & \textbf{Avg. Size} & \textbf{GPops/s} \\
            \hline
            100       & 0.08    & 402.12 & $2.84 \times 10^{9}$ & 0.19    & 400.63 & $8.45 \times 10^{9}$ & 4.19    & 376.45 & $1.88 \times 10^{10}$ & 6.71     & 336.03 & $5.20 \times 10^{10}$ \\
            500       & 0.32    & 473.41 & $4.49 \times 10^{9}$ & 0.83    & 482.12 & $1.19 \times 10^{10}$ & 13.81   & 430.74 & $4.29 \times 10^{10}$ & 45.00    & 434.43 & $5.69 \times 10^{10}$ \\
            1,000     & 0.55    & 454.55 & $4.98 \times 10^{9}$ & 1.44    & 460.11 & $1.35 \times 10^{10}$ & 55.20   & 474.11 & $2.07 \times 10^{10}$ & 116.24   & 454.86 & $4.23 \times 10^{10}$ \\
            5,000     & 2.78    & 479.52 & $5.21 \times 10^{9}$ & 8.71    & 487.41 & $1.12 \times 10^{10}$ & 292.27  & 486.24 & $2.07 \times 10^{10}$ & 575.76   & 465.43 & $4.63 \times 10^{10}$ \\
            10,000    & 6.66    & 486.38 & $4.41 \times 10^{9}$ & 17.83   & 483.29 & $1.07 \times 10^{10}$ & 547.45  & 468.78 & $1.85 \times 10^{10}$ & 910.28   & 480.79 & $5.38 \times 10^{10}$ \\
            50,000    & 33.61   & 475.78 & $4.27 \times 10^{9}$ & 98.74   & 482.26 & $9.65 \times 10^{9}$ & 3041.50 & 489.39 & $1.69 \times 10^{10}$ & 4960.14  & 463.82 & $4.84 \times 10^{10}$ \\
            100,000   & 77.99   & 474.46 & $3.66 \times 10^{9}$ & 223.00  & 488.90 & $8.65 \times 10^{9}$ & 7187.06 & 485.39 & $1.44 \times 10^{10}$ & 11626.46 & 458.33 & $4.17 \times 10^{10}$ \\
            500,000   & 892.91  & 467.66 & $1.79 \times 10^{9}$ & 2184.56 & 488.62 & $4.74 \times 10^{9}$ & --      & --     & --                   & --       & --     & --                   \\
            1,000,000 & 1098.94 & 464.58 & $2.54 \times 10^{9}$ & 2448.10 & 487.54 & $7.90 \times 10^{9}$ & --      & --     & --                   & --       & --     & --                   \\
            \hline
        \end{tabular}
    \end{table}
}

\newcommand{\tabEvoGPBreakdown}{
    \begin{table}[h]
        \centering
        \setlength{\tabcolsep}{2.5pt}
        \renewcommand{\arraystretch}{1.3}
        \caption{Detailed Breakdown of EvoGP GPops/s Performance Across Population and Dataset Sizes}
        \begin{tabular}{|r|c c c|c c c|c c c|c c c|}
            \hline
            \multirow{2}{*}[-1pt]{\makecell[r]{\textbf{Population} \\ \textbf{Size}}} & 
            \multicolumn{3}{c|}{\textbf{60 Datapoints}} & 
            \multicolumn{3}{c|}{\textbf{392 Datapoints}} & 
            \multicolumn{3}{c|}{\textbf{20640 Datapoints}} & 
            \multicolumn{3}{c|}{\textbf{100000 Datapoints}} \\
            \cline{2-13}
            & \textbf{Time} & \textbf{Avg. Size} & \textbf{GPops/s} 
            & \textbf{Time} & \textbf{Avg. Size} & \textbf{GPops/s} 
            & \textbf{Time} & \textbf{Avg. Size} & \textbf{GPops/s} 
            & \textbf{Time} & \textbf{Avg. Size} & \textbf{GPops/s} \\
            \hline
            100       & 0.91  & 309.30 & $2.05 \times 10^{8}$ & 0.89   & 293.35 & $1.26 \times 10^{9}$ & 1.06    & 171.64 & $3.49 \times 10^{10}$ & 1.45     & 234.81 & $1.55 \times 10^{11}$ \\
            500       & 1.07  & 392.24 & $1.20 \times 10^{9}$ & 1.05   & 460.96 & $8.67 \times 10^{9}$ & 2.40    & 291.49 & $1.27 \times 10^{11}$ & 4.48     & 310.63 & $3.35 \times 10^{11}$ \\
            1,000     & 1.38  & 409.62 & $1.85 \times 10^{9}$ & 1.87   & 461.65 & $1.02 \times 10^{10}$ & 4.06    & 327.67 & $1.61 \times 10^{11}$ & 7.66     & 287.79 & $3.50 \times 10^{11}$ \\
            5,000     & 1.48  & 344.26 & $7.31 \times 10^{9}$ & 3.16   & 467.77 & $2.93 \times 10^{10}$ & 21.21   & 424.70 & $2.07 \times 10^{11}$ & 52.43    & 434.40 & $4.14 \times 10^{11}$ \\
            10,000    & 1.84  & 336.85 & $1.13 \times 10^{10}$ & 4.33   & 473.37 & $4.32 \times 10^{10}$ & 41.63   & 426.26 & $2.08 \times 10^{11}$ & 112.52   & 439.56 & $3.92 \times 10^{11}$ \\
            50,000    & 12.00 & 421.12 & $1.67 \times 10^{10}$ & 30.52  & 486.44 & $3.38 \times 10^{10}$ & 228.26  & 470.74 & $2.14 \times 10^{11}$ & 562.32   & 460.16 & $4.10 \times 10^{11}$ \\
            100,000   & 11.91 & 427.72 & $2.27 \times 10^{10}$ & 36.00  & 484.57 & $5.31 \times 10^{10}$ & 458.80  & 470.49 & $2.12 \times 10^{11}$ & 1155.62  & 467.02 & $4.05 \times 10^{11}$ \\
            500,000   & 34.62 & 374.19 & $3.25 \times 10^{10}$ & 163.11 & 487.93 & $5.87 \times 10^{10}$ & 2465.43 & 474.55 & $1.99 \times 10^{11}$ & 6188.84  & 466.23 & $3.78 \times 10^{11}$ \\
            1,000,000 & 74.96 & 409.84 & $3.29 \times 10^{10}$ & 327.49 & 490.13 & $5.87 \times 10^{10}$ & 5002.68 & 476.15 & $1.97 \times 10^{11}$ & 11668.21 & 454.52 & $3.90 \times 10^{11}$ \\
            \hline
        \end{tabular}
    \end{table}
}

\newcommand{\tabTaskPerformance}{
    \begin{table*}[hbp]
    \centering
    \caption{Performance and Computational Time Across Tasks and Population Sizes.}
    \label{tab:task_performance}
    \begin{threeparttable}
    \renewcommand{\arraystretch}{1.3} 
    \setlength{\tabcolsep}{5pt} 
    \begin{tabular}{|c|c|c|c|c|c|}
    \hline
    \textbf{Task Type} & \textbf{Generations} & \textbf{Task Identifier} & \textbf{Pop Size} & \textbf{Performance} &  \textbf{Wall-clock Time (s)} \\ 
    \hline
    \multirow{12}{*}{Symbolic Regression} & \multirow{12}{*}{900} & \multirow{3}{*}{Daily Demand Forecasting} 
        & 200 & 377.004 $\pm$ 734.034 & 1.890 $\pm$ 0.067 \\
       & & & 1000 & 12.809 $\pm$ 10.681 & 3.928 $\pm$ 0.104 \\
       & & & 5000 & \textbf{1.400 $\pm$ 3.311} & 7.383 $\pm$ 0.463 \\
    \cline{3-6}
    & & \multirow{3}{*}{Auto MPG} 
        & 200 & 11.025 $\pm$ 3.517 & 2.209 $\pm$ 0.047 \\
    &    & & 1000 & 8.130 $\pm$ 2.515 & 5.771 $\pm$ 0.174 \\
    &    & & 5000 & \textbf{5.031 $\pm$ 0.812} & 16.725 $\pm$ 0.315 \\
    \cline{3-6}
    & & \multirow{3}{*}{California Housing} 
        & 200 & 0.535 $\pm$ 0.060 & 13.736 $\pm$ 0.978 \\
    &    & & 1000 & 0.408 $\pm$ 0.085 & 47.688 $\pm$ 9.969 \\
    &    & & 5000 & \textbf{0.354 $\pm$ 0.054} & 224.483 $\pm$ 7.479 \\
    \cline{3-6}
    & & \multirow{3}{*}{Feynman I.9.18} 
        & 200 & 2.253e-3 $\pm$ 1.727e-3 & 24.540 $\pm$ 2.785 \\
    &    & & 1000 & 6.961e-4 $\pm$ 7.430e-4 & 98.195 $\pm$ 7.242 \\
    &    & & 5000 & \textbf{2.783e-4 $\pm$ 2.816e-4} & 473.305 $\pm$ 18.692 \\
    \hline
    \multirow{12}{*}{Classification} & \multirow{12}{*}{300} & \multirow{3}{*}{Wine} 
        & 200 & 0.843 $\pm$ 0.109 & 60.061 $\pm$ 1.297 \\
    &     & & 1000 & 0.956 $\pm$ 0.013 & 70.997 $\pm$ 1.276 \\
    &     & & 5000 & \textbf{0.981 $\pm$ 0.008} & 98.401 $\pm$ 7.476 \\
    \cline{3-6}
    &  & \multirow{3}{*}{Chronic Kidney Disease} 
        & 200 & 0.943 $\pm$ 0.013 & 60.336 $\pm$ 1.961 \\
    &     & & 1000 & 0.963 $\pm$ 0.015 & 68.899 $\pm$ 3.310 \\
    &     & & 5000 & \textbf{0.973 $\pm$ 0.010} & 105.580 $\pm$ 5.803 \\
    \cline{3-6}
    &  & \multirow{3}{*}{Breast Cancer Wisconsin} 
        & 200 & 0.960 $\pm$ 0.009 & 63.457 $\pm$ 3.817 \\
    &     & & 1000 & 0.973 $\pm$ 0.005 & 69.717 $\pm$ 0.014 \\
    &     & & 5000 & \textbf{0.978 $\pm$ 0.005} & 122.362 $\pm$ 5.190 \\
    \cline{3-6}
    &  & \multirow{3}{*}{Yeast} 
        & 200 & 0.507 $\pm$ 0.026 & 62.417 $\pm$ 1.732 \\
    &     & & 1000 & 0.553 $\pm$ 0.033 & 81.128 $\pm$ 6.065 \\
    &     & & 5000 & \textbf{0.596 $\pm$ 0.010} & 214.529 $\pm$ 13.796 \\
    \hline
    \multirow{12}{*}{Robotics Control} &\multirow{12}{*}{100} & \multirow{3}{*}{Swimmer} 
        & 200 & 369.509 $\pm$ 1.414 & 791.546 $\pm$ 99.421 \\
    &     & & 1000 & 373.577 $\pm$ 2.363 & 891.185 $\pm$ 65.661 \\
    &     & & 5000 & \textbf{376.662 $\pm$ 2.585} & 735.877 $\pm$ 57.907 \\
    \cline{3-6}
    &  & \multirow{3}{*}{Hopper} 
        & 200 & 1399.618 $\pm$ 448.981 & 354.702 $\pm$ 25.832 \\
    &     & & 1000 & 2444.291 $\pm$ 464.516 & 386.617 $\pm$ 21.714 \\
    &     & & 5000 & \textbf{2748.185 $\pm$ 505.279} & 503.509 $\pm$ 2.800 \\
    \cline{3-6}
    &  & \multirow{3}{*}{Walker2d} 
        & 200 & 1550.955 $\pm$ 334.275 & 324.897 $\pm$ 62.063 \\
    &     & & 1000 & 1905.647 $\pm$ 189.443 & 553.908 $\pm$ 19.456 \\
    &     & & 5000 & \textbf{2550.197 $\pm$ 385.302} & 1233.198 $\pm$ 9.948 \\
    \cline{3-6}
    &  & \multirow{3}{*}{Halfcheetah} 
        & 200 & 5545.219 $\pm$ 716.671 & 1175.703 $\pm$ 77.945 \\
    &     & & 1000 & 11313.934 $\pm$ 3181.319 & 1458.846 $\pm$ 7.558 \\
    &     & & 5000 & \textbf{14114.018 $\pm$ 3136.102} & 3701.693 $\pm$ 22.334 \\
    \hline
    \end{tabular}
    \begin{tablenotes}
        \item[*] The best performances are highlighted in bold.
        \item[*] Performance metrics: (1) {Symbolic Regression}: MSE (lower is better); (2) {Classification}: Accuracy (higher is better); (3) {Robotics Control}: Cumulative Reward (higher is better).
    \end{tablenotes}
    \end{threeparttable}
    \end{table*}
}

\begin{document}

\title{
\fontsize{23}{29}\selectfont Enabling Population-Level Parallelism in Tree-Based Genetic Programming for GPU Acceleration
}

\author{Zhihong Wu,~\IEEEmembership{Student Member, IEEE,}
        Lishuang Wang,~\IEEEmembership{Student Member, IEEE,}
        Kebin Sun,~\IEEEmembership{Graduate Student Member, IEEE,}
        Zhuozhao Li,~\IEEEmembership{Member, IEEE}
        and Ran Cheng,~\IEEEmembership{Senior Member, IEEE}
    \thanks{The first three authors contributed equally to this work.}
    \thanks{This work was supported in part by Guangdong Basic and Applied Basic Research Foundation (No. 2024B1515020019).}
    \thanks{
    Zhihong Wu, Lishuang Wang, and Zhuozhao Li are with the Department of Computer Science and Engineering, Southern University of Science and Technology, Shenzhen 518055, China, and Kebin Sun was with the same department. 
    E-mails: \{zhihong2718, wanglishuang22, sunkebin.cn\}@gmail.com, lizz@sustech.edu.cn.
    }
    \thanks{Ran Cheng is with the Department of Data Science and Artificial Intelligence, and the Department of Computing, The Hong Kong Polytechnic University, Hong Kong SAR, China. He is also with The Hong Kong Polytechnic University Shenzhen Research Institute, Shenzhen, China. (E-mail: ranchengcn@gmail.com) (\emph{Corresponding author: Ran Cheng}).}
}


\maketitle

\begin{abstract}
Tree-based Genetic Programming (TGP) is a widely used evolutionary algorithm for tasks such as symbolic regression, classification, and robotic control. Due to the intensive computational demands of running TGP, GPU acceleration is crucial for achieving scalable performance. 
However, efficient GPU-based execution of TGP remains challenging, primarily due to three core issues: 
(1) the structural heterogeneity of program individuals, 
(2) the complexity of integrating multiple levels of parallelism, 
and (3) the incompatibility between high-performance CUDA execution and flexible Python-based environments.
To address these issues, we propose EvoGP, a high-performance framework tailored for GPU acceleration of TGP via population-level parallel execution. 
First, EvoGP introduces a tensorized representation that encodes variable-sized trees into fixed-shape, memory-aligned arrays, enabling uniform memory access and parallel computation across diverse individuals. 
Second, EvoGP adopts an adaptive parallelism strategy that dynamically combines intra- and inter-individual parallelism based on dataset size, ensuring high GPU utilization across a broad spectrum of tasks. 
Third, EvoGP embeds custom CUDA kernels into the PyTorch runtime, achieving seamless integration with Python-based environments such as Gym, MuJoCo, Brax, and Genesis.
\edit{
Experimental results demonstrate that EvoGP achieves a peak throughput exceeding $10^{11}$ GPops/s. Specifically, this performance represents a speedup of up to $304\times$ over existing GPU-based TGP implementations and $18\times$ over state-of-the-art CPU-based libraries. Furthermore, EvoGP maintains comparable accuracy and exhibits improved scalability across large population sizes.
EvoGP is open source and accessible at \url{https://github.com/EMI-Group/evogp}.
}

\end{abstract}


\section{Introduction}

\IEEEPARstart{G}{enetic} Programming (GP)~\cite{koza1994genetic} is an evolutionary algorithm that autonomously evolves computer programs based on the principles of natural selection and genetic inheritance. In contrast to black-box models such as neural networks, GP produces human-interpretable solutions and allows the structure of the program to evolve dynamically, rather than being fixed in advance. This capability to simultaneously search over both structural and parametric spaces reduces manual design effort and mitigates model bias, making GP a highly flexible and interpretable machine learning paradigm~\cite{mei2022explainable,de2025kozax}.
Since its inception in the early 1990s~\cite{koza1990paradigm}, GP has undergone extensive development, leading to a variety of variants that differ in their program representations and manipulation strategies. Based on structural representation, GP can be broadly classified into Linear GP~\cite{banzhaf1998lgp,brameier2007lgp}, Cartesian GP~\cite{miller1999cgp,miller2015cgp}, Gene Expression Programming~\cite{ferreira2001gep}, and Grammatical Evolution~\cite{o2001ge}. Among these, Tree-based Genetic Programming (TGP) has emerged as the most widely adopted variant~\cite{ari2021application_review,white2013community}, striking a favorable balance between evolutionary search efficiency and interpretability.
In TGP, programs are encoded as hierarchical tree structures, where internal nodes represent operators and leaf nodes denote variables or constants. This tree-based representation facilitates flexible and expressive modeling, and enables efficient evolutionary search via genetic operators such as crossover and mutation. As a result, TGP has been successfully applied to a wide range of tasks, including symbolic regression~\cite{schmidt2009distilling,chen2017sr,la2021contemporary_sr,makke2024sr_review}, classification~\cite{tran2016classification,espejo2009survey_classification,bi2021feature_learning}, complex system design~\cite{hornby2003robot,koza2005genetic4,lohn2008antenna,koza2010human}, and robotic control~\cite{perkins2000evolving_complex,dracopoulos2013solver,hein2018interpretable}.

While TGP has been successfully applied across a variety of domains, most existing implementations remain constrained to CPU-based execution~\cite{fortin2012deap, stephens2015gplearn, burlacu2020operon}. 
Although CPUs may suffice for small-scale problems, their performance rapidly degrades as data volume and task complexity increase. 
In modern applications, evaluating individuals often involves processing large datasets or simulating interactions within complex environments, i.e., a procedure that must be repeated for every member of the evolving population. 
This substantial computational overhead presents a major barrier to algorithmic refinement, particularly in tuning critical hyperparameters. 
It also severely restricts the adoption of meta-optimization strategies, such as automatic parameter tuning and adaptive operator selection, which have been shown to significantly enhance algorithm performance~\cite{huang2020auto-tuning, ma2025meta-black-box, chen2025meta-de}. 
These strategies typically require executing a large number of configurations, which becomes infeasible under high computational cost.
Consequently, the excessive evaluation time not only renders TGP impractical for large-scale or complex problems, but also hinders its evolution and adaptability through higher-level optimization mechanisms.
This substantially limits the practical utility of TGP in real-world scenarios where scalability, responsiveness, and adaptability are crucial.

To meet the growing computational demands, increasing efforts have been devoted to leveraging modern hardware architectures (e.g., GPUs) which offer massive parallelism and high throughput.
Specifically, two intuitive parallelization strategies have emerged to harness GPU capabilities for TGP.
The first strategy, known as intra-individual parallelism, involves evaluating a single individual across all data points in parallel. This approach is particularly well-suited for tasks such as symbolic regression and classification on large datasets, drawing inspiration from batch processing paradigms in deep learning. Most existing GPU-accelerated TGP implementations adopt this model, as it aligns naturally with the Single Instruction, Multiple Data (SIMD) execution model of GPU hardware~\cite{staats2017karoogp, baeta2021tensorgp, zhang2022srgpu}.
The second strategy, known as inter-individual parallelism, capitalizes on the inherent parallelism of evolutionary algorithms by simultaneously evaluating multiple candidate solutions.
{
This form of population-level parallelism has been shown to be effective and has become a core component of modern high-performance evolutionary computation. 
On one hand, it forms the basis of frameworks built on deep learning libraries such as JAX (e.g., EvoJAX~\cite{tang2022evojax}, evosax~\cite{lange2023evosax}) and PyTorch (e.g., EvoTorch~\cite{toklu2023evotorch}). 
On the other hand, similar strategies have been adopted in distributed general-purpose frameworks such as EvoX~\cite{huang2024evox} and in domain-specific acceleration, including neuroevolution~\cite{wang2024tensorneat} and evolutionary multiobjective optimization~\cite{liang2024tensorrvea, liang2025TensorMO}. 
However, despite its adoption in these areas, population-level parallelism remains relatively unexplored in TGP. 
}
When applied to TGP, inter-individual parallelism enables concurrent execution of both fitness evaluations and genetic operations (e.g., crossover, mutation, and selection), significantly reducing computational overhead. 
More importantly, it lays a solid foundation for scaling TGP to much larger populations, thereby enhancing both efficiency and exploratory capacity.

However, efficiently parallelizing TGP at the population level on GPUs remains fundamentally challenging due to several critical technical obstacles.
First, the most fundamental difficulty lies in the inherent structural heterogeneity of TGP individuals. 
Unlike tensor-based models, where computational structures are uniform and well-aligned for parallel execution, TGP individuals can vary significantly in size, topology, and evaluation logic. 
This structural diversity renders mainstream tensor computation frameworks such as TensorFlow~\cite{abadi2016tensorflow} and PyTorch~\cite{paszke2017pytorch} ill-suited for TGP. 
Even with the fine-grained control provided by low-level frameworks like CUDA, efficient GPU execution requires encoding these variable structures into a regularized and memory-aligned format~\cite{guide2020cuda}, which is a problem that remains largely unsolved.
Second, although data-level parallelism aligns naturally with the SIMD architecture of GPUs and has been effectively exploited in prior work, integrating it with inter-individual (population-level) parallelism introduces substantial system complexity. 
When both forms of parallelism are considered, the system must coordinate memory layouts, optimize kernel fusion, and manage execution scheduling to prevent resource contention and ensure high throughput. 
These trade-offs pose significant challenges in maintaining computational efficiency.
Third, many practical TGP applications are deployed in Python-based ecosystems such as OpenAI Gym~\cite{greg2016gym}, MuJoCo~\cite{todorov2012mujoco}, Brax~\cite{freeman2021brax}, and Genesis~\cite{2024genesis}. 
For a GPU-accelerated TGP system to be usable in these contexts, it must provide seamless Python interoperability without sacrificing performance. Achieving this balance between execution speed, development flexibility, and system extensibility constitutes a nontrivial engineering challenge.

To address the above challenges, we present EvoGP, a high-performance framework tailored for efficient and fully parallel population-level execution of TGP on GPUs.
EvoGP tackles the fundamental barriers of structural heterogeneity, multi-dimensional parallelism, and system integration through the following key contributions:

\begin{enumerate}
    \item We propose a tensorized representation for population-level parallelism on GPUs, which transforms structurally diverse TGP individuals into fixed-size, memory-aligned arrays and enables consistent memory access for high-throughput execution. 
    Based on this design, EvoGP is the first framework to achieve GPU acceleration across all evolutionary stages, including initialization, evaluation, and genetic operations.
    
    \item We develop an adaptive parallelism strategy for fitness evaluation, which dynamically combines intra-individual and inter-individual execution based on dataset size. 
    This approach allows EvoGP to maintain high GPU utilization and consistently efficient evaluation across tasks with varying input scales.
    
    \item We enhance the flexibility and extensibility of EvoGP through high-performance integration with Python environments. 
    By embedding custom CUDA operators into the PyTorch runtime, EvoGP enables seamless deployment in platforms such as Gym, MuJoCo, Brax, and Genesis, while offering rich features and customizable interfaces for research and application development.
    
    \item \edit{
    We empirically evaluate the proposed EvoGP.
    Extensive experiments demonstrate that EvoGP attains a peak throughput exceeding $10^{11}$~GPops/s, achieves speedups of up to $304\times$ over GPU-based TGP implementations and $18\times$ over the fastest CPU-based libraries, and improves scalability for large population sizes.
    }
\end{enumerate}

The remainder of this paper is structured as follows.  
Section~\ref{sec:background} introduces the fundamentals of TGP, describes the GPU computing model, and discusses existing TGP implementations.
Section~\ref{method} presents our tensorization strategy and adaptive evaluation scheme that enable scalable and efficient GPU-based execution of TGP.
Section~\ref{EvoGP_Implementations} details the implementation of the EvoGP system, including its integration with Python platforms and support for extensibility and multi-output programs.  
Section~\ref{experiments} reports the experimental results that validate the effectiveness and efficiency of our approach.  
Finally, Section~\ref{conclusion} concludes the paper.

\section{Background} \label{sec:background}
In this section, we provide an overview of the key background concepts relevant to this work. 
We begin by briefly reviewing the fundamentals of TGP, followed by an introduction to GPU architecture and the CUDA programming model. 
We then summarize representative existing TGP implementations, highlighting their capabilities and limitations in the context of parallel execution.

\subsection{Tree-based Genetic Programming}

TGP is a core evolutionary algorithm with broad applicability across several domains. 
In symbolic regression, TGP is widely used to discover mathematical expressions from data, effectively modeling complex relationships between variables~\cite{schmidt2009distilling,chen2017sr,la2021contemporary_sr,makke2024sr_review}. For classification tasks, TGP evolves both feature extraction methods and classification rules, facilitating the development of interpretable models~\cite{tran2016classification,espejo2009survey_classification,bi2021feature_learning}. In complex system design, TGP has been successfully used to automate the creation and optimization of electronic circuits, antennas, and other sophisticated systems~\cite{hornby2003robot,koza2005genetic4,lohn2008antenna,koza2010human}. Furthermore, TGP has shown effectiveness in robotic control by automatically generating robust control strategies for dynamic and uncertain environments~\cite{perkins2000evolving_complex,dracopoulos2013solver,hein2018interpretable}.

{In TGP, candidate solutions are represented as tree structures that explicitly define the computational flow from input to output. Formally, a TGP tree can be described as}
$$
T = (V, E, r),
$$
where $V$ is the set of nodes, $E \subseteq V \times V$ is the set of edges, and $r \in V$ is the distinguished root node. Every node $v \in V \setminus \{r\}$ has exactly one parent.

A TGP tree represents a function $\text{Func}$ that maps an input vector $\mathbf{x}$ to an output $y$:
$$
y = \text{Func}(\mathbf{x}).
$$
The computation proceeds in a bottom-up manner: each node $v$ calculates its result $res[v]$, and the root node's result $res[r]$ serves as the final output $y$.
Each node in the tree is represented by a tuple $v = (t,\, val)$, where $t$ is the node type, and $val$ stores the node’s value. Nodes in a TGP tree fall into three categories:

\noindent\textbf{Constant Node}: This represents a numerical constant. Its computation is:
$$
res[v] = val.
$$

\noindent\textbf{Variable Node}: This represents an input variable. Here, $val$ indicates the index of the input vector $\mathbf{x}$, and its computation is:
$$
res[v] = \mathbf{x}[val].
$$

\noindent\textbf{Function Node}: This represents a mathematical or logical function. In this case, $val$ specifies the function type $f$, and its computation is:
$$
res[v] = f\bigl(res[c_1],\, res[c_2],\, \dots,\, res[c_n]\bigr),
$$
where $c_1, c_2, \dots, c_n$ are the children of the node.

By using this tree-based representation, TGP supports flexible modeling of mathematical expressions and decision processes, which constitute the foundation of Genetic Programming. Fig.~\ref{fig:sigmoid_tree} shows an example of TGP trees.

\figSigmoidTree

\algTBGP

TGP iteratively refines a population of trees through a series of genetic operations, including selection, crossover, and mutation, to evolve optimal or near-optimal solutions.
Selection is a process that chooses individuals from the population based on their performance, ensuring that better-performing solutions have a higher probability of propagating to subsequent generations. Crossover involves exchanging information between selected parent trees, facilitating genetic diversity and exploration of the solution space. Mutation introduces random modifications by altering nodes or branches within an individual tree, thus promoting further diversity and preventing premature convergence~\cite{poli2008fieldguide}. The evolutionary process is described in Algorithm~\ref{alg:TBGP}.

\subsection{GPU Architecture and CUDA Programming}

With the increasing demand for parallel computing, GPUs have evolved into highly efficient accelerators for data-intensive tasks.
Modern GPUs consist of thousands of processing cores organized into Streaming Multiprocessors (SMs), enabling massive parallelism. The memory hierarchy includes high-latency global memory, multilevel caches (L1, L2) for faster access, and constant memory optimized for read-only data~\cite{pharr2005gpu_gems}. 
However, due to issues such as thread contention, synchronization overhead and fragmentation, dynamic memory management on GPUs exhibits significantly slower performance compared to CPUs. 
This makes handling variable-sized data structures, such as trees, particularly challenging~\cite{winter2021dynamic}.

Compute Unified Device Architecture (CUDA) is a parallel computing platform and programming model developed by NVIDIA, designed to harness the computing power of GPUs for general purpose computing~\cite{nickolls2008scalable_cuda,sanders2010cuda_gpgpu}. One of the fundamental execution models in CUDA is SIMT~\cite{lindholm2008SIMT}. CUDA organizes parallel execution in a hierarchical manner, consisting of three key levels: grid, block, and thread. A CUDA kernel launches a grid of thread blocks, where each block contains multiple threads. These threads execute in parallel, and the execution is coordinated through shared memory, synchronization mechanisms, and optimizations of the memory hierarchy~\cite{guide2020cuda,garland2008cuda_parallel}.

\subsection{Existing TGP Implementations} \label{sec:related_work}

Several TGP libraries have been developed in the past decade, exhibiting varying levels of computational efficiency, scalability, and extensibility.

{
DEAP~\cite{fortin2012deap} and gplearn~\cite{stephens2015gplearn} are representative CPU-based libraries for genetic programming. 
DEAP is designed for extensibility and supports a wide range of evolutionary algorithms with customizable operators and multiprocessing, whereas gplearn focuses on symbolic regression and classification and employs joblib to parallelize evaluations across CPU cores. 
In contrast, Operon~\cite{burlacu2020operon} is a modern CPU-based library that incorporates advanced parallel computing techniques to improve computational efficiency. 
\edit{
 Its C++ implementation exploits population-level parallelism through Taskflow~\cite{huang2021taskflow} and data-level parallelism through SIMD vectorization, which provides notable acceleration compared with earlier CPU-based GP libraries.
}
}

To improve computational efficiency, several libraries have sought to take advantage of GPU acceleration. KarooGP~\cite{staats2017karoogp} and TensorGP~\cite{baeta2021tensorgp} are built on the TensorFlow framework. KarooGP constructs static computation graphs for each GP individual and executes them using TensorFlow’s graph execution model. This design incurs substantial overhead due to repeated compilation, significantly slowing down execution. TensorGP adopts eager execution to avoid compilation, but still lacks fine-grained control over GPU resources. Combined with Python-level overhead and TensorFlow's abstraction cost, the resulting performance remains suboptimal.

SRGPU~\cite{zhang2022srgpu} instead implements core operations directly in CUDA for higher performance. Its design allows GPU threads to process all data points for a single individual in parallel, efficiently exploiting data-level parallelism. However, operations such as evaluation, selection, crossover, and mutation are still executed sequentially at the population level, leaving the majority of GPU concurrency untapped and limiting scalability. Furthermore, since SRGPU lacks Python bindings, it cannot be easily integrated into mainstream machine learning or robotics pipelines, which significantly limits its extensibility beyond symbolic regression tasks.

In addition, most existing libraries represent GP trees using traditional pointer-based or chain-like structures. While such representations are sufficient for serial execution within individuals, they are fundamentally unsuited for efficient population-level parallelism on GPUs. Their irregular and non-contiguous memory layout prevents memory coalescing and hinders the use of fast shared memory, resulting in inefficient global memory access and degraded throughput when scaling parallel execution across individuals.

\section{Methodology} \label{method}
In this section, we present the methodology underlying EvoGP for enabling efficient population-level parallelism in TGP for GPU acceleration. 
Specifically, we detail the key techniques that address the structural heterogeneity of individuals, support scalable genetic operations, and dynamically optimize GPU utilization.

\subsection{Tensorized Data Structures} 
\label{sec:tensorized_encoding}

To enable efficient population-level parallelism on GPUs, it is important to consider tree representations that allow for coalesced memory access and uniform tensor shapes. As discussed in Section~\ref{sec:related_work}, traditional pointer-based or chain-like structures are inadequate for this setting due to their irregular memory layout and poor compatibility with batched GPU execution.

To overcome these limitations, we adopt a linear prefix encoding scheme that represents each tree as a flat sequence of nodes. In general, prefix sequences alone are insufficient to fully represent arbitrary tree structures, as the number of children per node may vary. However, this ambiguity can be resolved in TGP, where the number of children for each node is entirely determined by its type ($t$) and its value ($val$). For instance, \textit{Constant} and \textit{Variable} nodes have no children, whereas a \textit{Unary Function} node (e.g., $\sin$) has one child, and a \textit{Binary Function} node (e.g., $\text{sum}$) has two children. Consequently, for TGP trees, the prefix sequence of node types and values can encode all critical information about the topology and values of a tree. This allows for a concise, linear representation of arbitrarily structured trees.

Based on this observation, we encode each tree $T$ using two arrays first:
{
\begin{align*}
    \mathbf{n}_{\text{type}} &= [t_1, t_2, \ldots, t_n] \in \mathbb{N}^{\text{len}(T)}, \\
    \mathbf{n}_{\text{val}} &= [val_1, val_2, \ldots, val_n] \in \mathbb{R}^{\text{len}(T)},
\end{align*}
}
where $t_i$ and $val_i$ represent the type and value of the $i$-th node, respectively, and $\text{len}(T)$ denotes the total number of nodes in the tree $T$. Although this representation is compact, it does not store explicit structural information. As a result, genetic operations such as subtree crossover and mutation typically require repeatedly parsing the prefix sequence to determine subtree boundaries, which incurs a time complexity of $\mathcal{O}(n)$ and leads to inefficient GPU execution.

To remedy this, we introduce an additional array that records the size of each subtree:
{
$$
\mathbf{n}_{\text{size}} = [\text{size}(1), \text{size}(2), \ldots, \text{size}(n)] \in \mathbb{N}^{\text{len}(T)},
$$
}
where $\text{size}(i)$ indicates the size of the subtree rooted at the $i$-th node. With this information, the boundaries of subtrees can be directly accessed in $\mathcal{O}(1)$ time, effectively eliminating the need for costly structural parsing and significantly improving the efficiency of GPU-based genetic operations.

Although prefix encoding provides a suitable representation for trees in TGP, it remains challenging to efficiently batch-process these structures on GPUs due to their variable sizes. To address this, we introduce a hyperparameter $|T|_\text{max}$, which denotes the maximum allowed number of nodes in a tree. Each of the arrays $\mathbf{n}_{\text{type}}$, $\mathbf{n}_{\text{val}}$, and $\mathbf{n}_{\text{size}}$ is then padded with \texttt{NaN} values to form fixed-length arrays:
{
\begin{align*}
    \hat{\mathbf{n}}_{\text{type}} &= [t_1, t_2, \ldots, \texttt{NaN}, \ldots, \texttt{NaN}] \in \mathbb{N}^{|T|_\text{max}}, \\
    \hat{\mathbf{n}}_{\text{val}} &= [val_1, val_2, \ldots, \texttt{NaN}, \ldots, \texttt{NaN}] \in \mathbb{R}^{|T|_\text{max}}, \\
    \hat{\mathbf{n}}_{\text{size}} &= [\text{size}(1), \text{size}(2), \ldots, \texttt{NaN}, \ldots, \texttt{NaN}] \in \mathbb{N}^{|T|_\text{max}}.
\end{align*}
}
Here, the actual tree data occupy the first $\text{len}(T)$ positions, and the remaining entries are filled with \texttt{NaN}. This padding strategy ensures that all trees share a consistent tensor shape, enabling straightforward GPU-friendly batching.

Once each tree is converted to padded arrays $\hat{\mathbf{n}}_{\text{type}}$, $\hat{\mathbf{n}}_{\text{val}}$, and $\hat{\mathbf{n}}_{\text{size}}$, we concatenate them across the population dimension. Let $P$ be the total number of trees in the population. Then, the population can be represented using three tensors:
{
\begin{align*}
    P_{\text{type}} &= \big[\hat{\mathbf{n}}_{\text{type}}^1, \hat{\mathbf{n}}_{\text{type}}^2, \ldots\big] \in \mathbb{N}^{P \times |T|_\text{max}}, \\
    P_{\text{val}} &= \big[\hat{\mathbf{n}}_{\text{val}}^1, \hat{\mathbf{n}}_{\text{val}}^2, \ldots\big] \in \mathbb{R}^{P \times |T|_\text{max}}, \\
    P_{\text{size}} &= \big[\hat{\mathbf{n}}_{\text{size}}^1, \hat{\mathbf{n}}_{\text{size}}^2, \ldots\big] \in \mathbb{N}^{P \times |T|_\text{max}},
\end{align*}
}
where $\hat{\mathbf{n}}_{\text{type}}^i$, $\hat{\mathbf{n}}_{\text{val}}^i$, and $\hat{\mathbf{n}}_{\text{size}}^i$ are the padded arrays corresponding to the $i$-th tree. By unifying all trees into these fixed-shape tensors, genetic operators and fitness evaluations can be designed to leverage GPU parallelism effectively.

Fig.~\ref{fig:pop_encoding} provides a graphical overview of this tensorized encoding scheme. In summary, storing the subtree size array alongside prefix-encoded values enables rapid access to individual subtrees, while padding to a fixed length circumvents the overhead of dynamic memory allocation and irregular indexing on GPUs. As a result, this approach paves the way for an efficient and scalable implementation of TGP that fully exploits modern GPU architectures.

\figPopEncoding

\subsection{Tensorized Operations}
\label{sec:genetic_ops}

The tensorized tree encoding introduced in Section~\ref{sec:tensorized_encoding} provides not only the basis for efficient batched evaluation, but also a unified structural representation that facilitates high-performance genetic operations on the GPU. 
Using this encoding, EvoGP supports crossover and mutation entirely on-device, enabling full GPU acceleration throughout the entire evolutionary process.

A key observation is that many widely used genetic operators in tree-based GP, such as one-point crossover, subtree mutation, and hoist mutation, share a common structural primitive: replacing a subtree at a given node with another. This motivates a generalized formulation referred to as \textit{subtree exchange}, which captures the essential transformation underlying a broad class of genetic operations.

Formally, we define the subtree exchange operation as
\[
\text{exchange}(T_{\text{old}}, k, T_{\text{new}}) \rightarrow T^{*},
\]
where \(T_{\text{old}}\) is the original tree for the subtree exchange, \(k\) denotes the root of the target subtree in \(T_{\text{old}}\), and \(T_{\text{new}}\) is the subtree that will replace the original subtree rooted at \(k\). Under our tensorized encoding, each tree is represented by three tensors: \(\mathbf{n_{\text{type}}}\), \(\mathbf{n_{\text{value}}}\), and \(\mathbf{n_{\text{size}}}\). We use \(\oplus\) to denote the tensor concatenation operation, and define
{
\begin{align*}
\mathbf{n}_{\text{type}}^{*} &= \mathbf{n}_{\text{type}}^{\text{old}}[1, 2, \dots, s-1] 
    \;\oplus\; \mathbf{n}_{\text{type}}^{\text{new}} 
    \;\oplus\; \mathbf{n}_{\text{type}}^{\text{old}}[e,e+1, \dots], \\
\mathbf{n}_{\text{value}}^{*} &= \mathbf{n}_{\text{value}}^{\text{old}}[1, 2, \dots, s-1] 
    \;\oplus\; \mathbf{n}_{\text{value}}^{\text{new}} 
    \;\oplus\; \mathbf{n}_{\text{value}}^{\text{old}}[e,e+1, \dots], \\
\mathbf{n}_{\text{size}}^{*} &= \mathbf{n}_{\text{size}}^{\text{updated}}[1, 2, \dots, s-1] 
    \;\oplus\; \mathbf{n}_{\text{size}}^{\text{new}} 
    \;\oplus\; \mathbf{n}_{\text{size}}^{\text{old}}[e,e+1, \dots],
\end{align*}
\edit{
where \(s = \text{index}(k)\) and \(e = s + \mathbf{n}_{\text{size}}^{\text{old}}[k]\) denote the starting and ending positions, respectively, of the original subtree in \(T_{\text{old}}\).
} 
The tensor \(\mathbf{n_{\text{size}}^{\text{updated}}}\) is computed by adding \(\ \Delta n=\mathbf{n_{\text{size}}^{\text{new}}}[1] - \mathbf{n}_{\text{size}}^{\text{old}}[k]\) to the subtree sizes of all ancestor nodes of \(k\):
\[
\mathbf{n}_{\text{size}}^{\text{updated}}[i] = 
\begin{cases}
\mathbf{n}_{\text{size}}^{\text{old}}[i] + \Delta n, & \text{if $i$ is an ancestor of $k$}, \\
\mathbf{n}_{\text{size}}^{\text{old}}[i], & \text{otherwise}.
\end{cases}
\]
In particular, if \(\mathbf{n}_{\text{size}}^{\text{old}}[1] + \Delta n\) exceeds the predefined maximum size constraint,} the subtree exchange operation will be considered invalid and the original tree \(T_{old}\) will be returned directly. This check is made efficient by maintaining the explicit subtree size array in our tensorized representation.

By focusing on subtree exchange, we can efficiently implement virtually all TGP genetic operations. For instance, the one-point crossover, a fundamental crossover variant, replaces a random subtree of one parent with a random subtree from the other parent:
\[
T^{*} = \text{exchange}(T_{1}, k, T_{2}^{\text{sub}}),
\]
where \(T_{1}\) and \(T_{2}\) are parent trees, \(k\) is a random node in \(T_{1}\), and \(T_{2}^{\text{sub}}\) is a randomly selected subtree from \(T_{2}\). 

Similarly, subtree mutation replaces a randomly selected subtree with a newly generated subtree:
\[
T^{*} = \text{exchange}(T, k, T_{\text{new}}),
\]
where \(T\) is the tree to be mutated, \(k\) is a random node in \(T\), and \(T_{\text{new}}\) denotes the newly generated subtree. Fig.~\ref{fig:ea_operation} provides a graphical overview of these operations and their corresponding transformations in the tensorized encoding.

Once subtree exchange is implemented as a GPU primitive, many genetic operators can be realized by simply changing how subtrees are selected or constructed. Since the core tensor manipulation logic remains unchanged, new operator variants can be added without modifying the GPU kernel itself. This design streamlines the development of new algorithms and ensures that all genetic operations benefit from high-throughput GPU execution. As a result, the method delivers both high performance and broad extensibility across genetic operators.

\figEaOperation

\subsection{Adaptive Parallelism in Fitness Evaluation}
\label{sec:adaptive_eval}

Symbolic regression (SR)~\cite{schmidt2009distilling,chen2017sr,la2021contemporary_sr,makke2024sr_review} is a central application of TGP, where each individual's fitness is computed by comparing its predicted outputs against target values across all data points in a dataset. To ensure high GPU utilization across datasets of varying sizes, EvoGP introduces an \textit{adaptive parallelism} strategy that dynamically selects the appropriate execution mode at runtime.

When the number of data points $D$ is small to moderate, evaluating a single individual alone does not fully saturate the GPU. To avoid under-utilization, EvoGP adopts a hybrid parallelism strategy that combines data-level and population-level parallelism within a single kernel launch. Specifically, it configures a two-dimensional CUDA grid of shape $(P,\, \text{block\_cnt})$, where $P$ is the population size and each block processes a segment of data points for a given tree:
\begin{align*}
\text{block\_size} &= \min(D, 1024),\\
\text{block\_cnt} &= \left\lceil \frac{D}{\text{block\_size}} \right\rceil.
\end{align*}

\figCudaBlocks

In this configuration, multiple CUDA blocks are assigned to evaluate each individual, with each block independently computing a partial fitness result over its assigned data segment. These partial results are subsequently aggregated using hardware-supported atomic operations to ensure a correct and efficient fitness accumulation. This hybrid parallelization strategy improves GPU utilization even when $D$ is relatively small, as inter-individual parallelism compensates for the limited workload per individual. Additionally, performing the entire evaluation in a single kernel launch minimizes the launch overhead and further improves the computational efficiency. Fig.~\ref{fig:cuda_blocks} (top right) illustrates this mode.

However, as the dataset size increases, the evaluation of a single individual on $D$ data points may already saturate all available CUDA cores. In such cases, adding population-level parallelism offers little to no additional benefit in GPU utilization. EvoGP therefore switches to a pure data-level parallelism strategy, where each individual's evaluation is launched in a separate kernel invocation. The purpose of this shift is to optimize memory access by loading the individual's tree into constant memory before the kernel launch. Since constant memory is immutable during execution but offers fast broadcast access across threads, this approach greatly accelerates memory reads during evaluation. Compared to the hybrid scheme, where trees are accessed from slower global memory, the constant memory design significantly improves memory throughput and overall runtime efficiency for large-scale workloads. Fig.~\ref{fig:cuda_blocks} (bottom right) illustrates this mode.

To support both cases without manual configuration, EvoGP integrates a runtime switching mechanism, referred to as \textit{Choose Parallelism Adaptively}, which selects the appropriate execution strategy based on the dataset size. The switching threshold is determined by comparing the number of data points $D$ with the total number of CUDA cores, {estimated from the number of Streaming Multiprocessors (SMs) and their count of cores per SM.} Section~\ref{experiment_hybrid} further examines how their performance varies with dataset size and, based on these results, investigates the trade-off between the two strategies.

Regardless of the parallelism scheme selected, the computation of a single individual's output follows the standard stack-based evaluation of a postfix (reverse Polish) expression. Although each tree is stored in a prefix-encoded format, the computation is carried out in reverse order, effectively simulating a postfix traversal. The total tree length is obtained from $\mathbf{n}_{\text{size}}[0]$, and nodes are processed from end to start. At each step, the node's arity is inferred from its type, and its output is computed by applying the corresponding operation to values on the evaluation stack. This execution pattern is naturally compatible with the prefix encoding format, making it particularly suitable for efficient implementation on GPUs.

\section{Implementation} \label{EvoGP_Implementations}
In this section, we present the overall system architecture of {EvoGP}, detailing its core design components and implementation features. 
{Specifically}, we describe how the framework integrates GPU-accelerated operations with high-level Python interfaces, and highlight several additional capabilities that enhance its usability, flexibility, and support for diverse application scenarios.

\subsection{Overview of EvoGP} \label{sec:overview_evogp}
The overall system architecture of EvoGP is depicted in Fig.\ref{fig:Structure}, which illustrates its modular components and their interactions. 
Specifically, EvoGP is implemented in Python and leverages the PyTorch framework~\cite{paszke2017pytorch} for efficient tensor manipulation and GPU resource management.

To optimize computational performance, EvoGP incorporates a suite of custom CUDA kernels for critical evolutionary operations, including tree initialization, mutation, crossover, inference, and fitness evaluation in symbolic regression (SR) tasks. 
To accommodate input datasets of varying sizes during SR evaluation, EvoGP adopts both data-parallel and hybrid-parallel execution strategies, as detailed in Section~\ref{sec:adaptive_eval}. 
These CUDA kernels are integrated into the PyTorch runtime as custom operators, enabling seamless compatibility with the broader PyTorch ecosystem.

\edit{
Complementing these kernel-level optimizations, EvoGP adopts a model for full-GPU residency, which ensures that data is not frequently transferred between the host and the device. By maintaining both the population and the evaluation context entirely on the GPU, EvoGP effectively eliminates the substantial communication overhead typical of traditional frameworks. This zero-copy design philosophy facilitates compatibility with modern reinforcement learning environments that are GPU-accelerated. Consequently, the architecture enables parallel simulations that are both efficient and end-to-end, thereby eliminating potential IO bottlenecks. Collectively, this design integrates the flexibility of Python-based development with the high-throughput capabilities of GPU acceleration, thereby ensuring that EvoGP is accessible and computationally efficient.
}

In addition to accelerating TGP algorithms, EvoGP includes built-in support for several representative application domains, including symbolic regression, policy optimization, and classification. It also provides a curated set of standardized benchmarks across these domains, facilitating systematic evaluation of different TGP variants and enabling fine-grained algorithmic tuning.

\figStructure

\subsection{Rich Features and User Customization}

EvoGP extends beyond basic GPU-accelerated TGP operations by offering a diverse set of evolutionary operator variants and a flexible interface for customization. {In addition to the problem benchmarks mentioned above, EvoGP provides multiple variations of key TGP operators, including selection, mutation, and crossover.} These variants allow users to freely combine and configure different evolutionary strategies, enabling the design of customized TGP algorithms tailored to specific optimization tasks. Table~\ref{table:gp_comparison} compares the range of operator variants implemented in the main TGP libraries, {highlighting that EvoGP provides the comprehensive set of supported operators.}

To further enhance usability, EvoGP offers an intuitive interface for defining custom operators and problem formulations. Users can leverage low-level CUDA implementations to efficiently develop and execute their own optimized genetic operators, ensuring high performance while maintaining flexibility. Additionally, EvoGP’s modular design allows researchers and practitioners to apply TGP algorithms to their own problem domains, making it a versatile tool for both academic research and real-world applications.

\subsection{Multi-Output Support}
EvoGP extends traditional TGP by incorporating multi-output functionality, allowing the application of TGP to tasks that require multiple output values, such as policy optimization in reinforcement learning. This enhancement enables EvoGP to handle complex problem domains where a single-output GP framework would be insufficient.

\figModiNodes

To achieve multi-output computation, {EvoGP uses Modi nodes~\cite{zhang2004multiple} in the tree structure.} These nodes possess two key characteristics:
\begin{enumerate}
    \item Each Modi node specifies the output to which its value contributes. The contributions from multiple Modi nodes associated with the same output are aggregated according to a predefined rule, which in this case is summation, to produce the final output value.
    \item If a Modi node has a parent node, it does not propagate its computed value upward like regular nodes. Instead, it directly adopts the value of its rightmost child node. This mechanism effectively transforms the tree structure into a directed acyclic graph (DAG), enabling the simultaneous computation of multiple outputs.
\end{enumerate}

This distinction between regular nodes and Modi nodes is crucial when performing multi-output computations. Although regular nodes operate under standard GP rules, Modi nodes follow a specialized processing scheme to ensure proper output assignment. This approach enables EvoGP to effectively support multi-output GP, making it well-suited for reinforcement learning and other tasks requiring multiple correlated outputs.
The computational logic of Modi nodes is illustrated in Fig.~\ref{fig:modi_nodes}, which provides a schematic view of how multiple outputs are generated and propagated within the EvoGP framework.

\tabGpComparison

\section{Experiments}~\label{experiments}
{
In this section, we evaluate the performance of EvoGP across a broad range of tasks and datasets. 
The experiments are designed to verify the effectiveness of the proposed GPU-accelerated framework in addressing the computational challenges inherent in TGP. 
Specifically, the evaluations aim to answer the following questions: 
(1) How do hybrid and data-level parallelism perform under different conditions, and how should the switching threshold be determined according to hardware and workload characteristics? 
(2) How does EvoGP compare with existing implementations in terms of execution performance and solution quality, while simultaneously validating its scalability with respect to population size? 
(3) How does EvoGP perform across different application domains, including symbolic regression, classification, and robotics control? 
Addressing these questions demonstrates the efficiency, scalability, and adaptability of EvoGP across diverse computational scenarios.
}

\subsection{Experimental Setup}~\label{experiment setup}

{This section presents the experimental configuration used to evaluate the performance of {EvoGP}. All experiments were executed on a high-performance workstation, whose hardware specifications are summarized in Table~\ref{tab:system_specifications}. Standard TGP parameter settings commonly adopted in the literature were employed. To rigorously assess performance, key hyperparameters (i.e., population size and number of data points) were systematically varied, whereas the remaining parameters were fixed as listed in Table~\ref{tab:sr_params}. Each experiment was independently repeated ten times with distinct random seeds to ensure statistical robustness. The reported results are given as mean values with 95\% confidence intervals, providing a consistent and reliable basis for evaluating the efficiency, scalability, and adaptability of {EvoGP}.
}

\tabSystemSpecifications

\tabSrParams

\subsection{Threshold Analysis for Adaptive Parallelism}
\label{experiment_hybrid}

To validate the effectiveness of EvoGP's adaptive parallelism strategy, we conducted an experiment to identify the optimal threshold for switching between hybrid and data-level parallelism. This analysis was performed using a symbolic regression task benchmarked on the {Pagie polynomial}, a widely used non-linear function in GP research defined as:
$$f(x, y) = \frac{1}{1+x^{-4}} + \frac{1}{1+y^{-4}}.$$

Fig.~\ref{fig:experiment2} presents the wall-clock execution time as a function of the number of data points (x-axis) for four population sizes: 100, 1{,}000, 5{,}000, and 10{,}000. The y-axis denotes the average execution time in seconds. Across all configurations, a consistent performance trend is observed: hybrid parallelism yields lower execution time when the dataset is small, while data-level parallelism becomes more efficient as the dataset grows. This crossover behavior reflects the GPU utilization characteristics described in Section~\ref{sec:adaptive_eval}. When the number of data points is insufficient to fully occupy the GPU via data-level execution alone, hybrid parallelism leverages population-level concurrency to fill idle cores. As the dataset size increases, evaluating a single individual becomes sufficient to saturate the device, making population-level parallelism redundant. In this case, data-level parallelism further benefits from constant memory access, which provides faster broadcast capabilities than the global memory used in hybrid execution.

Although $D = 16{,}384$ is the first experimental point at which data-level parallelism visibly outperforms hybrid parallelism, the actual crossover appears to occur closer to $D \approx 10{,}496$. This value coincides with the total number of CUDA cores on the NVIDIA GeForce RTX 3090, which has 82 streaming multiprocessors with 128 cores each, totaling $82 \times 128 = 10{,}496$. This alignment suggests that data-level parallelism becomes more effective once the dataset size is large enough to fully utilize GPU's computational resources without requiring additional population-level work.

These findings provide empirical support for EvoGP's runtime switching mechanism, which selects the parallelism strategy by comparing the dataset size against the total number of available CUDA cores. By grounding this decision in hardware-aware thresholds, EvoGP dynamically adapts to diverse computational workloads and consistently maintains high execution efficiency.

\figExperimentTwo

\subsection{Comparison with Existing TGP Implementations}
\label{experiment_comparison}

\edit{
To benchmark the performance of EvoGP, we selected several representative libraries for comparison. For CPU-based frameworks, we considered three established TGP libraries: DEAP \cite{fortin2012deap}, gplearn \cite{stephens2015gplearn}, and Operon \cite{burlacu2020operon}. Since Operon provides alternative mathematical backends that affect performance results, we utilized the Eve backend to enhance computational throughput. For GPU-based counterparts, we included SRGPU \cite{zhang2022srgpu}. Furthermore, PySR \cite{cranmer2023pysr} was incorporated as an additional reference. A supplementary comparison with the TensorGP GPU library \cite{baeta2021tensorgp} is provided in the Supplementary Document.
All benchmarks were performed on four symbolic regression tasks collected from diverse sources. These tasks include Daily Demand Forecasting and Auto MPG from the UCI Machine Learning Repository \cite{dua2019uci}, California Housing from scikit-learn \cite{pedregosa2011scikit}, and Feynman I.9.18 from the Feynman Symbolic Regression Database \cite{udrescu2020ai}. Detailed specifications of these tasks are summarized in Table \ref{tab:sr_datasets}. Each experiment was executed for 100 generations. To ensure a consistent comparative timeframe, we only present results for the runs that completed within one hour.
}

\tabSrDatasets

{
As presented in Fig.~\ref{fig:experiment1-1}, the experimental results provide a comprehensive comparison of execution time and solution quality across different population sizes and datasets.
The top row shows the total execution time as a function of population size and uses a logarithmic scale for each dataset. 
Across all scenarios EvoGP consistently exhibits markedly lower execution times. 
While baseline libraries display approximately linear scaling with population size and appear exponential due to the logarithmic axis, EvoGP demonstrates a flatter scaling curve. 
It maintains nearly constant execution time at smaller population sizes and shows only moderate increases for populations in the hundreds of thousands. 
The bottom row presents solution quality measured by mean squared error (MSE). 
EvoGP matches or surpasses the accuracy of competing libraries and confirms that its substantial computational speedup does not come at the expense of solution quality.
}

\figExperimentOneOne

\tabGpopsSummary

While wall-clock time provides a direct measure of speed, Langdon~\cite{langdon2010gpops} introduced a metric that more precisely reflects the operational throughput of GP systems, termed GP operations per second (GPops/s). 
\edit{
The GPops/s metric is defined as the total number of genetic program primitive operations evaluated per second across all fitness cases, calculated as:
\[
\text{GPops/s} = \frac{G \times P \times \bar{S} \times D}{T},
\]
where $G$ denotes the number of generations, $P$ the population size, $\bar{S}$ the average program size measured by the number of nodes, $D$ the number of individual fitness cases (i.e., data points), and $T$ the total execution time in seconds.
}

\figExperimentOneTwo

\figExperimentOneThree

\edit{
EvoGP achieves robust computational throughput, maintaining a magnitude of approximately $10^{10}$~GPops/s on small datasets and reaching $10^{11}$~GPops/s on medium-to-large datasets, as detailed in Table~\ref{tab:gpops_summary}. Based on this high throughput, we compared EvoGP against the leading GPU-based library, SRGPU, and the CPU-based library, Operon, as illustrated in Fig.~\ref{fig:experiment1-2} and Fig.~\ref{fig:experiment1-3}. The results indicate that EvoGP holds a substantial performance advantage. Specifically, on smaller problems, EvoGP achieves peak speedups of $304\times$ relative to SRGPU and $18\times$ relative to Operon. In scenarios involving medium to large datasets, data-point parallelism saturates GPU resources, which naturally constrains the speedup magnitude. Nevertheless, EvoGP maintains a robust lead, thereby delivering at least a $3\times$ speedup over SRGPU. Against Operon, EvoGP sustains an approximately $10\times$ advantage, a margin that becomes particularly pronounced when the population size exceeds $1{,}000$. Although the hardware generational gap favors the GPU (14nm CPU vs. 8nm GPU), the order-of-magnitude speedups achieved by EvoGP remain substantial. Consequently, even with an estimated 40\% gain from a more modern 7nm CPU architecture, the throughput advantage of EvoGP remains clearly evident.
}
A detailed breakdown of these results, including the exact GPops/s values for all compared libraries, is provided in the Supplementary Document.

{
Taken together, these results demonstrate the efficiency and robustness of EvoGP’s GPU-accelerated design. The observed performance gain results from the customized design of EvoGP, which utilizes population-level parallelism while sustaining efficient GPU utilization.
}

\subsection{Performance in Multiple Tasks with Scalable Populations}
\label{experiment_multitasks}

\edit{
We evaluate the performance of EvoGP across three domains: symbolic regression, classification, and robotics control. For all tasks, we test population sizes of 200, 1{,}000, and 5{,}000. In symbolic regression, the experiments utilize the benchmarks described in Section~\ref{experiment_comparison} and adopt MSE as the performance metric. Specifically, lower values indicate superior accuracy. \edit{In classification, we employ four datasets from the UCI repository~\cite{dua2019uci}: \textit{Wine}, \textit{Chronic Kidney Disease}, \textit{Breast Cancer Wisconsin}, and \textit{Yeast}.} These datasets exhibit heterogeneity in the number of input features, output classes, and data instances. Table~\ref{tab:classification_datasets} summarizes these specifications. We measure performance via classification accuracy, where higher values denote better results. In robotics control, we utilize four environments from the Brax framework~\cite{freeman2021brax}: \textit{Swimmer}, \textit{Hopper}, \textit{Walker2d}, and \textit{HalfCheetah}. Table~\ref{tab:brax_environment_dimensions} lists the observation and action dimensions of these environments. Performance is evaluated by the cumulative reward obtained during an episode. Consequently, higher rewards represent improved control effectiveness.
}

\tabClassificationDatasets

\tabBraxEnvironmentDimensions

\tabPerformanceSummaryWide

{
The results are shown in Fig.~\ref{fig:experiment3} and Table~\ref{tab:performance_summary_wide}. 
The figure presents task performance with respect to wall-clock time and illustrates the convergence behaviour for each population size, whereas summarizes the final performance of each population at the end of the corresponding runtime. 
A comparison of performance over a fixed number of generations is provided in the Supplementary Document.
}

\figExperimentThree

{
Across all three domains, the results indicate that larger populations yield lower MSE in regression, higher accuracy in classification, and higher cumulative rewards in control tasks. 
As shown in Table~\ref{tab:performance_summary_wide}, the improvement is most evident in the HalfCheetah control task, where increasing the population from 200 to 5{,}000 more than doubled the cumulative reward, and in the Daily Demand Forecasting regression task, where the final MSE decreased by over two orders of magnitude. 
\edit{
Similarly, in the Yeast classification task, increasing the population size enabled EvoGP to surpass the standard accuracy of Neural Networks ($56.87\%$) and closely approach the best accuracy reported in the UCI repository benchmarks ($62.80\%$, achieved by Random Forest). 
}
These observations suggest that larger populations may contribute to improved GP performance by enabling more comprehensive explorations of the solution space.
}

{
The key factor underlying the performance of large populations lies in EvoGP's GPU-accelerated population-level parallelism. 
This mechanism enables scalability without a proportional increase in computation time. 
As illustrated in Fig.~\ref{fig:experiment3}, the efficiency of this design allows larger populations to reach higher-quality solutions in less wall-clock time than smaller ones. 
This property enables EvoGP to outperform CPU-based or sequential implementations in both computational speed and final solution quality.
}

\section{Conclusion}~\label{conclusion}

In this paper, we presented EvoGP, a fully GPU-accelerated framework for Tree-based Genetic Programming (TGP) that effectively addresses three fundamental challenges: the structural heterogeneity of individuals, the complexity of combining multiple levels of parallelism, and the tension between high-performance execution and Python-based flexibility. EvoGP introduces a tensorized representation that transforms structurally diverse trees into fixed-shape, memory-aligned arrays, thereby enabling efficient and consistent memory access for population-level GPU execution. 

To further enhance evaluation efficiency, we proposed an adaptive parallelism strategy that dynamically fuses intra- and inter-individual parallelism based on dataset size. 
EvoGP seamlessly integrates with Python environments via custom CUDA operators within the PyTorch runtime, thereby allowing effortless deployment in popular platforms such as Gym and MuJoCo. 
\edit{
Experimental results demonstrate that EvoGP attains a peak throughput exceeding $10^{11}$~GPops/s, which corresponds to speedups of up to $304\times$ over GPU-based TGP implementations and $18\times$ over the fastest CPU-based libraries. Furthermore, EvoGP preserves solution quality and improves scalability for large populations.
}

Looking forward, we aim to extend EvoGP to support more complex and diverse applications. Its computational efficiency and scalable parallelism provide a foundation for utilizing larger and more diverse populations, which can enhance performance in demanding scenarios such as robotic control and autonomous decision-making. While the current implementation supports multi-output tasks, future research will explore more advanced strategies to further enhance expressiveness and generalizability. We envision EvoGP as a robust and extensible framework that empowers researchers to advance the frontiers of interpretable and evolutionary AI.

\bibliographystyle{IEEEtran}
\bibliography{EvoGP}

@article{koza1994genetic,
  title     = {Genetic programming as a means for programming computers by natural selection},
  author    = {Koza, John R},
  journal   = {Statistics and Computing},
  volume    = {4},
  pages     = {87--112},
  year      = {1994},
  publisher = {Springer}
}

@article{fortin2012deap,
author = {Fortin, F\'{e}lix-Antoine and De Rainville, Fran\c{c}ois-Michel and Gardner, Marc-Andr\'{e} Gardner and Parizeau, Marc and Gagn\'{e}, Christian},
title = {{DEAP}: evolutionary algorithms made easy},
year = {2012},
issue_date = {January 2012},
publisher = {JMLR.org},
volume = {13},
number = {1},
issn = {1532-4435},
journal = {J. Mach. Learn. Res.},
month = jul,
pages = {2171–2175},
numpages = {5}
}

@misc{stephens2015gplearn,
  author       = {T. Stephens},
  title        = {gplearn: Genetic programming in {Python}, with a scikit-learn inspired {API}},
  year         = {2015},
  howpublished = {https://github.com/trevorstephens/gplearn}
}

@inproceedings{burlacu2020operon,
    author = {Burlacu, Bogdan and Kronberger, Gabriel and Kommenda, Michael},
    title = {{Operon C++}: An Efficient Genetic Programming Framework for Symbolic Regression},
    year = {2020},
    isbn = {9781450371278},
    publisher = {Association for Computing Machinery},
    address = {New York, NY, USA},
    doi = {10.1145/3377929.3398099},
    booktitle = {Proceedings of the 2020 Genetic and Evolutionary Computation Conference Companion},
    pages = {1562–1570},
    numpages = {9},
    location = {Canc\'{u}n, Mexico},
    series = {GECCO '20}
}

@inproceedings{staats2017karoogp,
author = {Staats, Kai and Pantridge, Edward and Cavaglia, Marco and Milovanov, Iurii and Aniyan, Arun},
title = {{TensorFlow} enabled genetic programming},
year = {2017},
isbn = {9781450349390},
publisher = {Association for Computing Machinery},
address = {New York, NY, USA},
doi = {10.1145/3067695.3084216},
booktitle = {Proceedings of the Genetic and Evolutionary Computation Conference Companion},
pages = {1872–1879},
numpages = {8},
location = {Berlin, Germany},
series = {GECCO '17}
}

@inproceedings{baeta2021tensorgp,
	title = {{TensorGP} -- Genetic Programming Engine in {TensorFlow}},
	author = {Baeta, Francisco and Correia, Jo{\~{a}}o and Martins, Tiago and Machado, Penousal},
	booktitle = {Applications of Evolutionary Computation - 24th International Conference, EvoApplications 2021},
	year = {2021},
	organization={Springer}
}

@inproceedings{zhang2022srgpu,
  title        = {Speeding up genetic programming based symbolic regression using {GPUs}},
  author       = {Zhang, Rui and Lensen, Andrew and Sun, Yanan},
  booktitle    = {Pacific Rim International Conference on Artificial Intelligence},
  pages        = {519--533},
  year         = {2022},
  organization = {Springer}
}

@article{huang2024evox,
  title = {{EvoX}: {A} {Distributed} {GPU}-accelerated {Framework} for {Scalable} {Evolutionary} {Computation}},
  author = {Huang, Beichen and Cheng, Ran and Li, Zhuozhao and Jin, Yaochu and Tan, Kay Chen},
  journal = {IEEE Transactions on Evolutionary Computation},
  year = 2024,
  note={Early access},
  doi = {10.1109/TEVC.2024.3388550}
}

@inproceedings{zhang2004multiple,
  title        = {A multiple-output program tree structure in genetic programming},
  author       = {Zhang, Yun and Zhang, Mengjie},
  booktitle    = {Proceedings of The Second Asian-Pacific Workshop on Genetic Programming},
  year         = {2004},
  organization = {Citeseer}
}

@book{poli2008fieldguide,
  author    = {Poli, Riccardo and Langdon, William B. and McPhee, Nicholas Freitag},
  title     = {A Field Guide to Genetic Programming},
  year      = {2008},
  isbn      = {1409200736},
  publisher = {Lulu Enterprises, UK Ltd}
}

@inproceedings{tang2022evojax,
title     = {{EvoJAX}: Hardware-accelerated neuroevolution},
author    = {Tang, Yujin and Tian, Yingtao and Ha, David},
year = {2022},
isbn = {9781450392686},
publisher = {Association for Computing Machinery},
address = {New York, NY, USA},
doi = {10.1145/3520304.3528770},
booktitle = {Proceedings of the Genetic and Evolutionary Computation Conference Companion},
pages = {308–311},
numpages = {4},
location = {Boston, Massachusetts},
series = {GECCO '22}
}

@inproceedings{lange2023evosax,
title     = {evosax: {JAX-based} evolution strategies},
author    = {Lange, Robert Tjarko},
year = {2023},
isbn = {9798400701207},
publisher = {Association for Computing Machinery},
address = {New York, NY, USA},
doi = {10.1145/3583133.3590733},
booktitle = {Proceedings of the Companion Conference on Genetic and Evolutionary Computation},
pages = {659–662},
numpages = {4},
location = {Lisbon, Portugal},
series = {GECCO '23}
}

@article{wang2024tensorneat,
  author = {Wang, Lishuang and Zhao, Mengfei and Liu, Enyu and Sun, Kebin and Cheng, Ran},
  title = {TensorNEAT: A GPU-accelerated Library for NeuroEvolution of Augmenting Topologies},
  year = {2025},
  publisher = {Association for Computing Machinery},
  address = {New York, NY, USA},
  doi = {10.1145/3730406},
  journal = {ACM Transactions on Evolutionary Learning and Optimization},
  month = apr
}

@inproceedings{liang2024tensorrvea,
author = {Liang, Zhenyu and Jiang, Tao and Sun, Kebin and Cheng, Ran},
title = {{GPU-accelerated} Evolutionary Multiobjective Optimization Using Tensorized {RVEA}},
year = {2024},
isbn = {9798400704949},
publisher = {Association for Computing Machinery},
address = {New York, NY, USA},
doi = {10.1145/3638529.3654223},
booktitle = {Proceedings of the Genetic and Evolutionary Computation Conference},
pages = {566–575},
numpages = {10},
location = {Melbourne, VIC, Australia},
series = {GECCO '24}
}

@misc{toklu2023evotorch,
  title   = {{EvoTorch}: Scalable Evolutionary Computation in {Python}},
  author  = {Toklu, Nihat Engin and Atkinson, Timothy and Micka, Vojt\v{e}ch and Liskowski, Pawe\l{} and Srivastava, Rupesh Kumar},
  year={2023},
  eprint={2302.12600},
  archivePrefix={arXiv},
  primaryClass={cs.NE},
  url={https://arxiv.org/abs/2302.12600}, 
}

@inproceedings{miller1999cgp,
author = {Miller, Julian F.},
title = {An empirical study of the efficiency of learning boolean functions using a Cartesian Genetic Programming approach},
year = {1999},
isbn = {1558606114},
publisher = {Morgan Kaufmann Publishers Inc.},
address = {San Francisco, CA, USA},
booktitle = {Proceedings of the 1st Annual Conference on Genetic and Evolutionary Computation - Volume 2},
pages = {1135–1142},
numpages = {8},
location = {Orlando, Florida},
series = {GECCO'99}
}

@inproceedings{miller2015cgp,
author = {Miller, Julian and Turner, Andrew},
title = {Cartesian Genetic Programming},
year = {2015},
isbn = {9781450334884},
publisher = {Association for Computing Machinery},
address = {New York, NY, USA},
doi = {10.1145/2739482.2756571},
booktitle = {Proceedings of the Companion Publication of the 2015 Annual Conference on Genetic and Evolutionary Computation},
pages = {179–198},
numpages = {20},
location = {Madrid, Spain},
series = {GECCO Companion '15}
}

@book{banzhaf1998lgp,
  author    = {Banzhaf, Wolfgang and Francone, Frank D. and Keller, Robert E. and Nordin, Peter},
  title     = {Genetic programming: an introduction: on the automatic evolution of computer programs and its applications},
  year      = {1998},
  isbn      = {155860510X},
  publisher = {Morgan Kaufmann Publishers Inc.}
}

@book{brameier2007lgp,
  title     = {Basic concepts of linear genetic programming},
  author    = {Brameier, Markus F and Banzhaf, Wolfgang},
  year      = {2007},
  publisher = {Springer}
}

@article{ferreira2001gep,
  title={Gene Expression Programming: A New Adaptive Algorithm for Solving Problems},
  author={Ferreira, C{\^a}ndida},
  journal={Complex Systems},
  volume={13},
  number={2},
  pages={87--129},
  year={2001}
}

@article{o2001ge,
  title     = {Grammatical evolution},
  author    = {O'Neill, Michael and Ryan, Conor},
  journal   = {IEEE Transactions on Evolutionary Computation},
  volume    = {5},
  number    = {4},
  pages     = {349--358},
  year      = {2001},
  publisher = {IEEE}
}

@article{white2013community,
  title     = {Better {GP} benchmarks: community survey results and proposals},
  author    = {White, David R and McDermott, James and Castelli, Mauro and Manzoni, Luca and Goldman, Brian W and Kronberger, Gabriel and Ja{\'s}kowski, Wojciech and O’Reilly, Una-May and Luke, Sean},
  journal   = {Genetic Programming and Evolvable Machines},
  volume    = {14},
  pages     = {3--29},
  year      = {2013},
  publisher = {Springer}
}

@article{mei2022explainable,
  title     = {Explainable artificial intelligence by genetic programming: A survey},
  author    = {Mei, Yi and Chen, Qi and Lensen, Andrew and Xue, Bing and Zhang, Mengjie},
  journal   = {IEEE Transactions on Evolutionary Computation},
  volume    = {27},
  number    = {3},
  pages     = {621--641},
  year      = {2022},
  publisher = {IEEE}
}

@article{ari2021application_review,
  title     = {A review of genetic programming: Popular techniques, fundamental aspects, software tools and applications},
  author    = {Ar{\i}, Davut and Alag{\"o}z, Bar{\i}{\c{s}} Baykant},
  journal   = {Sakarya University Journal of Science},
  volume    = {25},
  number    = {2},
  pages     = {397--416},
  year      = {2021},
  publisher = {Sakarya University}
}

@article{chen2017sr,
  title     = {Feature selection to improve generalization of genetic programming for high-dimensional symbolic regression},
  author    = {Chen, Qi and Zhang, Mengjie and Xue, Bing},
  journal   = {IEEE Transactions on Evolutionary Computation},
  volume    = {21},
  number    = {5},
  pages     = {792--806},
  year      = {2017},
  publisher = {IEEE}
}

@article{schmidt2009distilling,
  title     = {Distilling free-form natural laws from experimental data},
  author    = {Schmidt, Michael and Lipson, Hod},
  journal   = {Science},
  volume    = {324},
  number    = {5923},
  pages     = {81--85},
  year      = {2009},
  publisher = {American Association for the Advancement of Science}
}

@article{la2021contemporary_sr,
  title     = {Contemporary symbolic regression methods and their relative performance},
  author    = {La Cava, William and Burlacu, Bogdan and Virgolin, Marco and Kommenda, Michael and Orzechowski, Patryk and de Fran{\c{c}}a, Fabr{\'\i}cio Olivetti and Jin, Ying and Moore, Jason H},
  journal   = {Advances in Neural Information Processing Systems},
  volume    = {2021},
  number    = {DB1},
  pages     = {1},
  year      = {2021},
  publisher = {NIH Public Access}
}

@article{makke2024sr_review,
  title     = {Interpretable scientific discovery with symbolic regression: a review},
  author    = {Makke, Nour and Chawla, Sanjay},
  journal   = {Artificial Intelligence Review},
  volume    = {57},
  number    = {1},
  pages     = {2},
  year      = {2024},
  publisher = {Springer}
}

@article{koza2010human,
  title     = {Human-competitive results produced by genetic programming},
  author    = {Koza, John R},
  journal   = {Genetic Programming and Evolvable Machines},
  volume    = {11},
  pages     = {251--284},
  year      = {2010},
  publisher = {Springer}
}

@book{koza2005genetic4,
  title     = {Genetic programming {IV}: Routine human-competitive machine intelligence},
  author    = {Koza, John R and Keane, Martin A and Streeter, Matthew J and Mydlowec, William and Yu, Jessen and Lanza, Guido},
  volume    = {5},
  year      = {2005},
  publisher = {Springer Science \& Business Media}
}

@article{hornby2003robot,
  title     = {Generative representations for the automated design of modular physical robots},
  author    = {Hornby, Gregory S and Lipson, Hod and Pollack, Jordan B},
  journal   = {IEEE Transactions on Robotics and Automation},
  volume    = {19},
  number    = {4},
  pages     = {703--719},
  year      = {2003},
  publisher = {IEEE}
}

@article{lohn2008antenna,
  title     = {Human-competitive evolved antennas},
  author    = {Lohn, Jason D and Hornby, Gregory S and Linden, Derek S},
  journal   = {AI EDAM},
  volume    = {22},
  number    = {3},
  pages     = {235--247},
  year      = {2008},
  publisher = {Cambridge University Press}
}

@article{tran2016classification,
  title     = {Genetic programming for feature construction and selection in classification on high-dimensional data},
  author    = {Tran, Binh and Xue, Bing and Zhang, Mengjie},
  journal   = {Memetic Computing},
  volume    = {8},
  pages     = {3--15},
  year      = {2016},
  publisher = {Springer}
}

@article{espejo2009survey_classification,
  title     = {A survey on the application of genetic programming to classification},
  author    = {Espejo, Pedro G and Ventura, Sebasti{\'a}n and Herrera, Francisco},
  journal   = {IEEE Transactions on Systems, Man, and Cybernetics, Part C (Applications and Reviews)},
  volume    = {40},
  number    = {2},
  pages     = {121--144},
  year      = {2009},
  publisher = {IEEE}
}

@book{bi2021feature_learning,
  title     = {Genetic programming for image classification: An automated approach to feature learning},
  author    = {Bi, Ying and Xue, Bing and Zhang, Mengjie},
  volume    = {24},
  year      = {2021},
  publisher = {Springer Nature}
}

@book{pharr2005gpu_gems,
  title     = {{GPU Gems 2}: Programming techniques for high-performance graphics and general-purpose computation},
  author    = {Pharr, Matt and Fernando, Randima},
  year      = {2005},
  publisher = {Addison-Wesley Professional}
}

@inproceedings{winter2021dynamic,
  title     = {Are dynamic memory managers on {GPUs} slow? a survey and benchmarks},
  author    = {Winter, Martin and Parger, Mathias and Mlakar, Daniel and Steinberger, Markus},
  booktitle = {Proceedings of the 26th ACM SIGPLAN Symposium on Principles and Practice of Parallel Programming},
  pages     = {219--233},
  year      = {2021}
}

@inproceedings{nickolls2008scalable_cuda,
  author    = {Nickolls, John and Buck, Ian and Garland, Michael and Skadron, Kevin},
  title     = {Scalable parallel programming with {CUDA}},
  year      = {2008},
  isbn      = {9781450378451},
  publisher = {Association for Computing Machinery},
  doi       = {10.1145/1401132.1401152},
  booktitle = {ACM SIGGRAPH 2008 Classes},
  articleno = {16},
  numpages  = {14},
  location  = {Los Angeles, California},
  series    = {SIGGRAPH '08}
}

@book{sanders2010cuda_gpgpu,
  title     = {{CUDA} by Example: An Introduction to General-Purpose {GPU} Programming},
  author    = {Sanders, Jason},
  year      = {2010},
  publisher = {Addison-Wesley Professional}
}

@article{guide2020cuda,
  title   = {{CUDA C++} programming guide},
  author  = {Guide, Design},
  journal = {NVIDIA, July},
  year    = {2020}
}

@article{garland2008cuda_parallel,
  title     = {Parallel computing experiences with {CUDA}},
  author    = {Garland, Michael and Le Grand, Scott and Nickolls, John and Anderson, Joshua and Hardwick, Jim and Morton, Scott and Phillips, Everett and Zhang, Yao and Volkov, Vasily},
  journal   = {IEEE Micro},
  volume    = {28},
  number    = {4},
  pages     = {13--27},
  year      = {2008},
  publisher = {IEEE}
}

@article{lindholm2008SIMT,
  title     = {{NVIDIA Tesla}: A unified graphics and computing architecture},
  author    = {Lindholm, Erik and Nickolls, John and Oberman, Stuart and Montrym, John},
  journal   = {IEEE Micro},
  volume    = {28},
  number    = {2},
  pages     = {39--55},
  year      = {2008},
  publisher = {IEEE}
}

@article{hein2018interpretable,
  title     = {Interpretable policies for reinforcement learning by genetic programming},
  author    = {Hein, Daniel and Udluft, Steffen and Runkler, Thomas A},
  journal   = {Engineering Applications of Artificial Intelligence},
  volume    = {76},
  pages     = {158--169},
  year      = {2018},
  publisher = {Elsevier}
}

@article{dracopoulos2013solver,
  title     = {Genetic programming as a solver to challenging reinforcement learning problems},
  author    = {Dracopoulos, Dimitris C and Effraimidis, Dimitrios and Nichols, Barry D},
  journal   = {International Journal of Computer Research},
  volume    = {20},
  number    = {3},
  pages     = {351},
  year      = {2013},
  publisher = {Nova Science Publishers, Inc.}
}

@incollection{perkins2000evolving_complex,
  title     = {Evolving complex visual behaviours using genetic programming and shaping},
  author    = {Perkins, Simon and Hayes, Gillian},
  booktitle = {Interdisciplinary Approaches to Robot Learning},
  pages     = {162--184},
  year      = {2000},
  publisher = {World Scientific}
}

@misc{abadi2016tensorflow,
  title   = {{TensorFlow}: Large-scale machine learning on heterogeneous distributed systems},
  author  = {Abadi, Mart{\'\i}n and Agarwal, Ashish and Barham, Paul and Brevdo, Eugene and Chen, Zhifeng and Citro, Craig and Corrado, Greg S and Davis, Andy and Dean, Jeffrey and Devin, Matthieu and others},
  year={2016},
  eprint={1603.04467},
  archivePrefix={arXiv},
  primaryClass={cs.DC},
  url={https://arxiv.org/abs/1603.04467}, 
}

@inproceedings{paszke2017pytorch,
  title     = {Automatic differentiation in {PyTorch}},
  author    = {Paszke, Adam and Gross, Sam and Chintala, Soumith and Chanan, Gregory and Yang, Edward and DeVito, Zachary and Lin, Zeming and Desmaison, Alban and Antiga, Luca and Lerer, Adam},
  booktitle = {NIPS-W},
  year      = {2017}
}

@misc{de2025kozax,
  title={{Kozax}: Flexible and Scalable Genetic Programming in {JAX}},
  author={de Vries, Sigur and Keemink, Sander W and van Gerven, Marcel AJ},
  year={2025},
  eprint={2502.03047},
  archivePrefix={arXiv},
  primaryClass={cs.NE},
  url={https://arxiv.org/abs/2502.03047}, 
}

@techreport{koza1990paradigm,
author = {Koza, John R.},
title = {Genetic programming: a paradigm for genetically breeding populations of computer programs to solve problems},
year = {1990},
institution = {Stanford University},
address = {Stanford, CA, USA},
}

@ARTICLE{ma2025meta-black-box,
  author={Ma, Zeyuan and Guo, Hongshu and Gong, Yue-Jiao and Zhang, Jun and Tan, Kay Chen},
  journal={IEEE Transactions on Evolutionary Computation}, 
  title={Toward Automated Algorithm Design: A Survey and Practical Guide to Meta-Black-Box-Optimization}, 
  year={2025},
  volume={},
  number={},
  doi={10.1109/TEVC.2025.3568053},
  note = {Early access}
}

@ARTICLE{chen2025meta-de,
  author={Chen, Minyang and Feng, Chenchen and Cheng, Ran},
  journal={IEEE Transactions on Evolutionary Computation}, 
  title={{MetaDE}: Evolving Differential Evolution by Differential Evolution}, 
  year={2025},
  volume={},
  number={},
  note={Early access},
  doi={10.1109/TEVC.2025.3541587}
}

@ARTICLE{huang2020auto-tuning,
  author={Huang, Changwu and Li, Yuanxiang and Yao, Xin},
  journal={IEEE Transactions on Evolutionary Computation}, 
  title={A Survey of Automatic Parameter Tuning Methods for Metaheuristics}, 
  year={2020},
  volume={24},
  number={2},
  pages={201-216},
  doi={10.1109/TEVC.2019.2921598}
}

@inproceedings{todorov2012mujoco,
  title={{MuJoCo}: A physics engine for model-based control},
  author={Todorov, Emanuel and Erez, Tom and Tassa, Yuval},
  booktitle={2012 IEEE/RSJ International Conference on Intelligent Robots and Systems},
  pages={5026--5033},
  year={2012},
  organization={IEEE},
  doi={10.1109/IROS.2012.6386109}
}

@misc{greg2016gym,
  Author = {Brockman, Greg and Cheung, Vicki and Pettersson, Ludwig and Schneider, Jonas and Schulman, John and Tang, Jie and Zaremba, Wojciech},
  Title = {{OpenAI Gym}},
  year={2016},
  eprint={1606.01540},
  archivePrefix={arXiv},
  primaryClass={cs.LG},
  url={https://arxiv.org/abs/1606.01540}, 
}

@misc{freeman2021brax,
  title   = {{Brax}--a differentiable physics engine for large scale rigid body simulation},
  author  = {Freeman, C Daniel and Frey, Erik and Raichuk, Anton and Girgin, Sertan and Mordatch, Igor and Bachem, Olivier},
      year={2021},
      eprint={2106.13281},
      archivePrefix={arXiv},
      primaryClass={cs.RO},
      url={https://arxiv.org/abs/2106.13281}, 
}

@misc{2024genesis,
  author = {{Genesis Authors}},
  title = {Genesis: A Generative and Universal Physics Engine for Robotics and Beyond},
  month = {December},
  year = {2024},
  url = {https://github.com/Genesis-Embodied-AI/Genesis}
}

@inproceedings{langdon2010gpops,
  title={A many threaded {CUDA} interpreter for genetic programming},
  author={Langdon, William B},
  booktitle={European Conference on Genetic Programming},
  pages={146--158},
  year={2010},
  organization={Springer}
}

@misc{cranmer2023pysr,
  title={Interpretable machine learning for science with {PySR} and SymbolicRegression. jl},
  author={Cranmer, Miles},
  year={2023},
  eprint={2305.01582},
  archivePrefix={arXiv},
  primaryClass={astro-ph.IM},
  url={https://arxiv.org/abs/2305.01582}, 
}

@misc{dua2019uci,
  author = {Dua, Dheeru and Graff, Casey},
  year = {2017},
  title = {{UCI} Machine Learning Repository},
  url = {http://archive.ics.uci.edu/ml},
  institution = {University of California, Irvine, School of Information and Computer Sciences}
}

@article{daily_demand_forecasting_orders_409,
  title={Study on Daily Demand Forecasting Orders using Artificial Neural Network},
  author={Ricardo Pinto Ferreira and Andr{\'e}a Martiniano and Arthur Arruda Leal Ferreira and Aleister Ferreira and Renato Jos{\'e} Sassi},
  journal={IEEE Latin America Transactions},
  year={2016},
  volume={14},
  pages={1519-1525},
}

@misc{auto_mpg_9,
  author       = {Quinlan, R.},
  title        = {{Auto MPG}},
  year         = {1993},
  howpublished = {UCI Machine Learning Repository},
  url         = {https://doi.org/10.24432/C5859H}
}

@article{wine_109,
  title={Comparative analysis of statistical pattern recognition methods in high dimensional settings},
  author={Stefan Aeberhard and Danny Coomans and Olivier Y. de Vel},
  journal={Pattern Recognit.},
  year={1994},
  volume={27},
  pages={1065-1077},
}

@misc{yeast_110,
  author       = {Nakai, Kenta},
  title        = {{Yeast}},
  year         = {1991},
  howpublished = {UCI Machine Learning Repository},
  url         = {https://doi.org/10.24432/C5KG68}
}

@ARTICLE{liang2025tensormo,
  author={Liang, Zhenyu and Li, Hao and Yu, Naiwei and Sun, Kebin and Cheng, Ran},
  journal={IEEE Transactions on Evolutionary Computation}, 
  title={Bridging Evolutionary Multiobjective Optimization and {GPU} Acceleration via Tensorization}, 
  year={2025},
  volume={},
  number={},
  note={Early access},
  doi={10.1109/TEVC.2025.3555605}
}

@article{huang2021taskflow,
  author = {Huang, Tsung-Wei and Lin, Dian-Lun and Lin, Chun-Xun and Lin, Yibo},
  title = {Taskflow: A Lightweight Parallel and Heterogeneous Task Graph Computing System},
  year = {2022},
  issue_date = {June 2022},
  publisher = {IEEE Press},
  volume = {33},
  number = {6},
  issn = {1045-9219},
  journal = {IEEE Transactions on Parallel and Distributed Systems},
  month = jun,
  pages = {1303–1320},
  numpages = {18}
}

@article{california_housing,
  title = {Sparse spatial autoregressions},
  journal = {Statistics \& Probability Letters},
  volume = {33},
  number = {3},
  pages = {291-297},
  year = {1997},
  issn = {0167-7152},
  author = {R. {Kelley Pace} and Ronald Barry},
  publisher={Elsevier}
}

@article{udrescu2020ai,
  title={{AI Feynman}: A physics-inspired method for symbolic regression},
  author={Udrescu, Silviu-Marian and Tegmark, Max},
  journal={Science Advances},
  volume={6},
  number={16},
  pages={eaay2631},
  year={2020},
  publisher={American Association for the Advancement of Science}
}

@misc{chronic_kidney_disease_336,
  author       = {Rubini, L. and Soundarapandian, P. and Eswaran, P.},
  title        = {{Chronic Kidney Disease}},
  year         = {2015},
  howpublished = {UCI Machine Learning Repository},
  note         = {{DOI}: https://doi.org/10.24432/C5G020}
}

@article{pedregosa2011scikit,
  title={Scikit-learn: Machine learning in {Python}},
  author={Pedregosa, Fabian and Varoquaux, Ga{\"e}l and Gramfort, Alexandre and Michel, Vincent and Thirion, Bertrand and Grisel, Olivier and Blondel, Mathieu and Prettenhofer, Peter and Weiss, Ron and Dubourg, Vincent and others},
  journal={the Journal of Machine Learning Research},
  volume={12},
  pages={2825--2830},
  year={2011},
  publisher={JMLR. org}
}

\onecolumn

\setcounter{page}{1}

\begin{@twocolumnfalse}
\begin{center}
    \fontsize{23}{29}\selectfont  Enabling Population-Level Parallelism in Tree-Based Genetic Programming for GPU Acceleration \\
    (Supplementary Document)
    \vspace{0.5em} 
\end{center}
\end{@twocolumnfalse}

\setcounter{algorithm}{0}
\setcounter{table}{0}
\setcounter{figure}{0}
\setcounter{section}{0}
\setcounter{equation}{0}
\renewcommand\thealgorithm{S.\arabic{algorithm}}
\renewcommand\thetable{S.\Roman{table}}
\renewcommand{\figurename}{\normalsize Fig.}
\renewcommand\thefigure{S.\arabic{figure}}
\renewcommand\thesection{S.\Roman{section}}
\renewcommand\theequation{S.\arabic{equation}}

\vspace{1em} 

\setcounter{section}{0}

\section{Performance Comparison with TensorGP on the Pagie Polynomial Task}
\label{appendix:tensorgp}

{
In the main body of this paper, we benchmarked EvoGP against several state-of-the-art symbolic regression libraries. During the preliminary evaluations, we also examined TensorGP~\cite{baeta2021tensorgp}, a library designed for GPU acceleration. However, its support for custom datasets was found to be limited, and the initial results indicated relatively lower performance. To maintain a consistent comparison with current state-of-the-art libraries, the results for TensorGP were therefore not included in the main body of the paper.
}

{
For completeness, we conducted an additional experiment on the Pagie polynomial task, which TensorGP supports natively. This experiment compared EvoGP, TensorGP, and other representative libraries (i.e., SRGPU, DEAP, and gplearn) under identical conditions to provide a direct performance assessment. The experimental environment and hyperparameter settings were consistent with those in Table~\ref{tab:system_specifications} and Table~\ref{tab:sr_params}, and all GPU-based tests were performed on a single NVIDIA GeForce RTX~3090 GPU. The results, presented in Fig.~\ref{fig:tensorgp}, indicate that TensorGP achieved lower throughput than other GPU-based libraries and also underperformed compared to CPU-based implementations.
}

\figTensorgp

\clearpage
\section{Reproducibility and Software Environment}
\label{sec:appendix_software}

To facilitate the reproducibility of our experiments and provide comprehensive details regarding the computational environment, we present detailed versioning information for all symbolic regression libraries utilized in this study. Furthermore, we include specific build configurations for the \textit{Operon} library to ensure transparency regarding compiler optimizations and backend selections.

\subsection{Software Versions}
Table~\ref{tab:software_versions} lists the specific versions or Git commit hashes for the Python packages and C++/CUDA libraries utilized in our benchmarks.

\begin{table}[h]
    \centering
    \caption{Software Libraries and Versioning Information Used in the Experiments.}
    \label{tab:software_versions}
    \renewcommand{\arraystretch}{1.2}
    \begin{tabular}{@{}llp{7cm}@{}}
        \toprule
        \textbf{Library} & \textbf{Type} & \textbf{Version / Commit Hash} \\
        \midrule
        DEAP & Python Package & v1.4.3 \\
        gplearn & Python Package & v0.4.2 \\
        PySR & Python Package & v1.5.9 \\
        \midrule
        EvoGP & Source Build & \texttt{606f7f4aba8d1237cf0c818fe59c24457115b773} \\
        SRGPU & Source Build & \texttt{e7f0eb80bc38c23df11a06244649a2453d0fae74} \\
        TensorGP & Source Build & \texttt{f14708cb68238553cf8f735928837c2311264ab5} \\
        Operon & Source Build & \texttt{d13d3ea826744c98287d0fd9df61f7af56d8d71a} \\
        \bottomrule
    \end{tabular}
\end{table}

\subsection{Operon Build Details}
The performance of \textit{Operon} depends substantially on the underlying mathematical backend and compiler optimizations. Consequently, we provide the runtime version information and the precise CMake command used to build the library to ensure reproducibility.

\subsubsection{Runtime Version Information}
The output of the \texttt{./build/cli/operon\_gp --version} command on our experimental platform is as follows:

\begin{lstlisting}[language=bash, caption={Operon Version Output Details.}]
    operon rev. d13d3ea Release Linux-6.12.35 x86_64, timestamp 2025-12-31T03:15:48Z
    single-precision build using eigen 5.0.1, ceres n/a, taskflow 3.11.0
    compiler: GNU 13.2.0, flags: -g -O3 -DNDEBUG -Wno-error
\end{lstlisting}

\subsubsection{CMake Build Command}
To enable the \texttt{Eve} math backend and thereby maximize performance on the target CPU hardware, we utilized the following CMake invocation:

\begin{lstlisting}[language=bash, caption={CMake Command Used to Compile Operon with the Eve Backend.}]
cmake -S . \
    -B build --preset build-linux \
    -DCMAKE_BUILD_TYPE=Release \
    -DMATH_BACKEND="Eve" \
    -DBUILD_CLI_PROGRAMS=ON \
    -DCMAKE_EXPORT_COMPILE_COMMANDS=ON \
    -DCMAKE_CXX_FLAGS="-Wno-error" \
    -DBUILD_TESTING=OFF \
    -DBUILD_SHARED_LIBS=OFF \
    -DCMAKE_POSITION_INDEPENDENT_CODE=ON \
    -DCMAKE_TOOLCHAIN_FILE=/root/vcpkg/scripts/buildsystems/vcpkg.cmake \
    -DVCPKG_OVERLAY_PORTS=./ports \
    -DCMAKE_PREFIX_PATH=/root/vcpkg/installed/x64-linux \
    -DCMAKE_C_COMPILER=/usr/bin/gcc \
    -DCMAKE_CXX_COMPILER=/usr/bin/c++
\end{lstlisting}
\color{black}

\clearpage
\section{GPops/s Performance Data for All Libraries}~\label{appendix:gpops}

\tabPySRBreakdown

\tabDEAPBreakdown

\tabgplearnBreakdown

\tabSRGPUBreakdown

\tabOperonBreakdown

\tabEvoGPBreakdown

\clearpage
\section{Performance Summary for the Fixed Generation}~\label{appendix:performance_summary}

\tabTaskPerformance

\end{document}